\documentclass[runningheads]{llncs}

\usepackage{eccv}

\usepackage{eccvabbrv}

\usepackage{graphicx}
\usepackage{booktabs}
\usepackage{bbding}
\usepackage{multirow}
\usepackage{appendix}
\usepackage{makecell}
\usepackage{authblk}

\usepackage[accsupp]{axessibility}  % Improves PDF readability for those with disabilities.

\usepackage[pagebackref,breaklinks,colorlinks,citecolor=eccvblue]{hyperref}
\usepackage{hyperref}

\usepackage{orcidlink}
\usepackage{array}
\newcolumntype{C}[1]{>{\centering}p{#1}}
\begin{document}

\title{HERM: Benchmarking and Enhancing Multimodal LLMs for  Human-Centric Understanding} 

\titlerunning{HERM}

% \newcommand*\equalnote{Three authors have equal contributions and are listed in alphabetical order.}
\author{Keliang Li*\inst{1} \and Zaifei Yang*\inst{1} \and Jiahe Zhao*\inst{1} \and Hongze Shen\inst{1} \and Ruibing Hou\inst{1} \and Hong Chang\inst{1} \and Shiguang Shan\inst{1} \and Xilin Chen\inst{1}}
\authorrunning{Li et al.}

\institute{Institute of Computing Technology, Chinese Academy of Sciences, China}

\maketitle

\begin{abstract}
% For the significant advances of Multimodal Large Language Models (MLLMs) in visual understanding and  instruction-following, diverse and common human-centric scenarios provide more possibilities for further applications.
The significant advancements in visual understanding and instruction following from Multimodal Large Language Models (MLLMs) have opened up more possibilities for broader applications in diverse and universal human-centric scenarios. However, existing image-text data may not support the precise modality alignment and integration of multi-grained information, which is crucial for human-centric visual understanding. In this paper, we introduce HERM-Bench, a benchmark for evaluating the human-centric understanding capabilities of MLLMs. Our work reveals the limitations of existing MLLMs in understanding complex human-centric scenarios. To address these challenges, we present HERM-100K, a comprehensive dataset with multi-level human-centric annotations, aimed at enhancing MLLMs' training. Furthermore, we develop HERM-7B, a MLLM that leverages enhanced training data from HERM-100K. Evaluations on HERM-Bench demonstrate that HERM-7B significantly outperforms existing MLLMs across various human-centric dimensions, reflecting the current inadequacy of data annotations used in MLLM training for human-centric visual understanding. This research emphasizes  the importance of specialized datasets and benchmarks in advancing the MLLMs' capabilities for human-centric understanding.
\end{abstract}

\section{Introduction}
\let\thefootnote\relax\footnotetext{*Three authors have equal contributions and are listed in alphabetical order.}
\label{sec:intro}
Benefiting from the remarkable breakthroughs of Large Language Models (LLMs)\cite{openai2023gpt,touvron2023llama,touvron2023llama2}, Multimodal Large Language Models (MLLMs)\cite{OpenAI2023GPT4v,team2023gemini,huang2024language,liu2024visual,zhu2023minigpt}, which equips LLMs with vision input, exhibit capabilities to perform open-ended visual understanding tasks\cite{liu2023improved,dai2023instructblip,chen2023minigpt}. Naturally, complex and frequently encountered \textit{human-centric scenarios} offer numerous potential applications for these advancements. 
Initial explorations have shown promise, including employing MLLMs for
reasoning about human roles \cite{zhang2023gpt4roi,liu2023mmbench, zhan2023generating}, predicting motion trajectories \cite{yu2023merlin, li2023seed, ning2023video} and grounding speakers in videos \cite{lin2023mm}. Additionally, the inherent open-ended capabilities of MLLMs position them as auxiliary tools \cite{lin2024motion, liu2023plan, xiao2023unified, brohan2023rt, zhao2022compositional} for various human-centric Artificial Intelligence Generative Content (AIGC) applications \cite{shi2023learning,tevet2022human}. Therefore, reliable visual understanding and complex-task execution  in human-centric scenarios enjoy an essential position for MLLMs.

\begin{figure}[tb]
  \centering
  \includegraphics[width=0.85\linewidth]{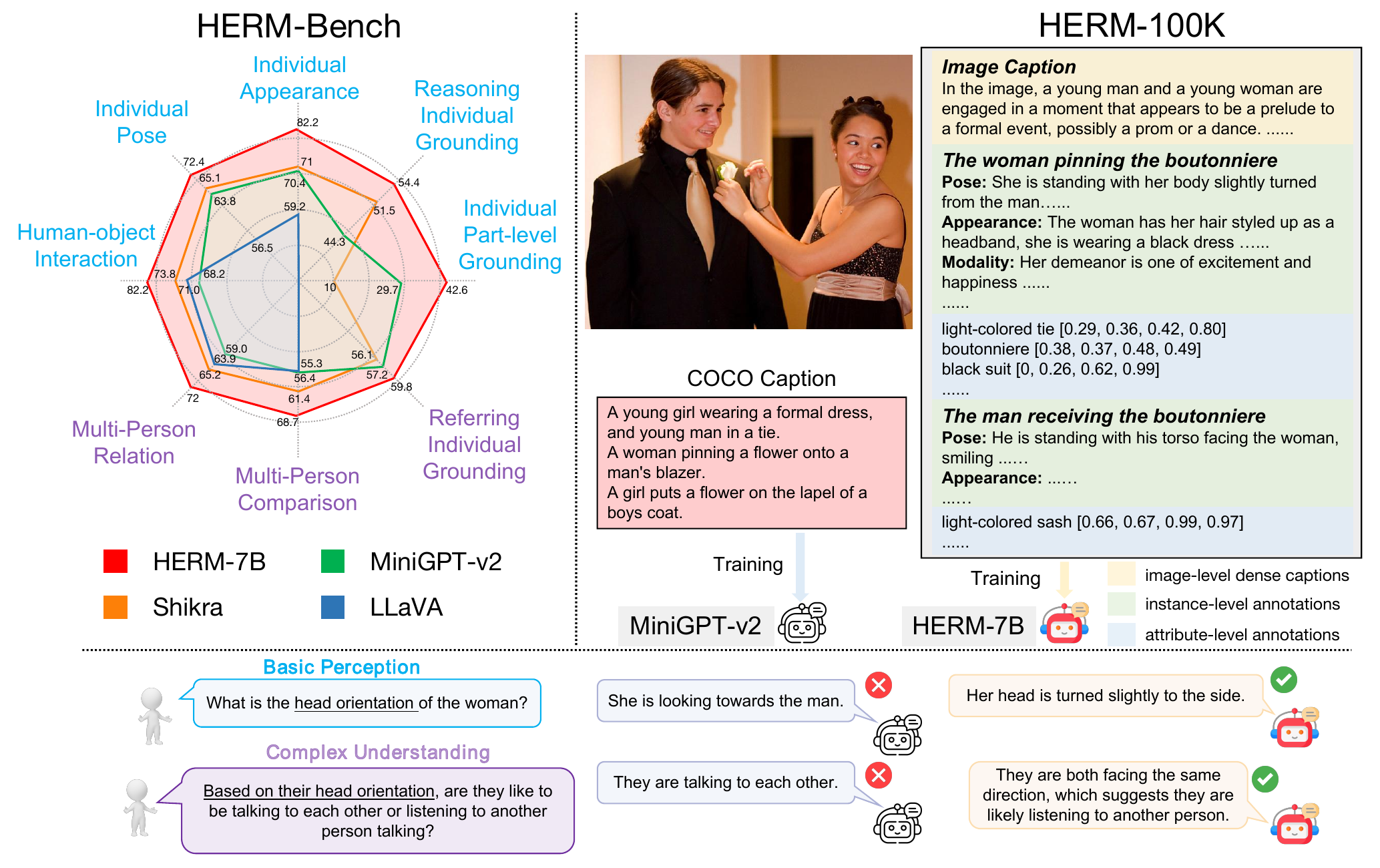}
    \vspace{-5pt}
  \caption{Overview of HERM. (1) We construct HERM-Bench, the first human-centric multi-modal benchmark. (2) We propose HERM-100K with multi-level human annotations. (3) We develop HERM-7B, a MLLM achieving state-of-the-art performance on human-centric basic perception and complex understanding.}
  \label{fig:teaser}
  \vspace{-20pt}
\end{figure}

Putting aside the diversity in model architecture, most MLLMs adopt a dual-phase paradigm encompassing a pre-training stage with large-scale image-text pairs for modality alignment, followed by a instruction fine-tuning stage that enhances multi-modal understanding capabilities through instruction-format data. Despite their achievements in general domain, \textbf{a concern is that the general visual understanding capability of MLLMs may not suffice for complex human-centric understanding}.  Previous human-centric perception generalists\cite{tang2023humanbench,ci2023unihcp} illustrate that all-side recognition of human needs to extract information in diverse data from multiple granularity. However, as shown in~\cref{fig:teaser} and~\cref{fig:review}, it is observed that captions from COCO\cite{lin2014microsoft}, widely used for training MLLMs, \textbf{fall short in scope and granularity} when describing humans, leading to drastically reduced human-related cues and details in text annotations.
As a consequence, MLLMs pre-trained and fine-tuned on these datasets may not achieve the desired performance under human-centric scenarios\cite{wang2023see,chen2023sharegpt4v,yu2023capsfusion}.

In light of this concern, in this work, we firstly focus on comprehensively evaluating the human-centric understanding capability of MLLMs, by introducing a benchmark named HERM{(\textbf{H}uman c\textbf{E}nt\textbf{R}ic \textbf{M}ulti-modality)}-Bench. HERM-Bench spans $8$ evaluation dimensions including \textit{basic perception} and \textit{complex understanding}, and comprises 2,748 questions involving multiple choice and grounding, as depicted in \cref{fig:teaser} and Fig.~\ref{fig:test-benchmark-taxonomy}. We design a sophisticated pipeline for generate the question-answer pairs tailed to evaluate specific dimensions. As shown in \cref{fig:teaser}, evaluations on HERM-Bench reveal that existing MLLMs exhibit severe limitations in human-centric perception and understanding scenarios.
Based on above analysis, we argue that the low-quality (fall short in scope and granularity) human annotations in existing datasets hinder the performance of MLLMs in human-centric visual understanding tasks.

To enhance human-centric understanding capability of MLLMs, we introduce HERM-100K, the first comprehensive  human-centric dataset for MLLM training. HERM-100K comprises over 100K human-centric annotations generated by GPT-4V \cite{OpenAI2023GPT4v}  with diverse image sources. As presented in~\cref{fig:teaser}, these annotations encompass \textit{multi-level} visual information, including \textit{image-level dense captions} capturing thorough scene details, \textit{instance-level annotations} covering multiple dimensions of humans, and \textit{attribute-level annotations} highlighting body parts and rare attributes. Through its multi-level structure, the annotations increase both the scope and granularity over raw captions, providing a comprehensive description of human-centric visual information.

Leveraging HERM-100K, we augment original training data with two components: 320K image/region-text pairs of captioning and grounding tasks constructed using pre-defined templates for multi-task pre-training stage \cite{dai2023instructblip,you2023ferret,chen2023shikra}; and 29K instruction-following pairs by prompting GPT-4\cite{openai2023gpt} based on our \textit{multi-level} annotations for instruction tuning stage \cite{liu2024visual}. Equipped with enhanced human-related annotations, we have developed a state-of-the-art large multi-modal model, HERM-7B. Despite without elaborate architecture design, our HERM-7B outperforms other MLLMs across all evaluation dimensions of HERM-Bench, showcasing its superiority in human-centric understanding. 

\section{Related Work}
In this section, we provide the related works about MLLMs. The related works about  human-centric foundation models are provided in Appendix.

\noindent{\textbf{Multimodal Large Language Models.}} \
Benefit from the success of LLMs~\cite{openai2023gpt,touvron2023llama,touvron2023llama2}, multi-modal models  \cite{alayrac2022flamingo,li2022blip,li2023blip,chen2023shikra}  achieve great improvements.
Recent MLLMs, such as PaLM-E~\cite{driess2023palm}, LLaVA~\cite{liu2024visual}, Shikra~\cite{chen2023shikra}, and MiniGPT-v2~\cite{chen2023minigpt},   leverage simple linear layer to bridge visual and language modality.
Furthermore, to enhance multimodal understanding capability, several studies focus on the quality of pretraining and finetuning datasets. 
For instance,  LLaVA~\cite{liu2024visual}, SVIT~\cite{zhao2023svit} and InstructBLIP~\cite{dai2023instructblip} enhance the quality of instruction-tuning data, advancing the comprehension of complex prompts. The works, including Shikra~\cite{chen2023shikra}, Ferret~\cite{you2023ferret} and KOSMOS-2~\cite{peng2023kosmos}, introduce new data types and training methods related to grounding, enhancing the grounding capability of MLLMs. 

Additionally, several studies~\cite{fan2024improving,lai2023scarcity,chen2023sharegpt4v} focus on enhancing the quality of captions within image-text pairs. For example, LaCLIP~\cite{fan2024improving} leverages LLMs to rewrite raw captions, yet its efficacy is limited by the low quality of raw captions.
The works~\cite{gadre2024datacomp,nguyen2024improving,lai2023scarcity} blend information from raw and synthetic captions. However, the caption fusion process overlooks the visual information, potentially leading to inevitable hallucination descriptions.    
ShareGPT4V~\cite{chen2023sharegpt4v} prompts GPT-4V to produce dense descriptions for images.  However, it primarily captures  global visual information for entire scene, potentially omitting detailed visual and locational information about specific person in the captions.
Different from \cite{chen2023sharegpt4v}, our work generates \textit{multi-level} annotations which can provide  fine-grained and comprehensive descriptions of person within the image,  thereby is more conducive to enhancing MLLMs' human-centric understanding capability.  

\noindent{\textbf{Benchmark for MLLMs.}} \
As MLLMs research progresses rapidly, some works~\cite{fu2023mme,liu2023mmbench,li2023seed,xu2023lvlm,bitton2024visit,you2023ferret} propose various comprehensive benchmarks for evaluating MLLMs. These benchmarks generally fall into three categories: multiple-choice questions~\cite{fu2023mme,li2023seed}; chat-based free-form output~\cite{xu2023lvlm,bitton2024visit}; localization tasks regarding referring and grounding~\cite{you2023ferret}. 
However, existing benchmarks focus on evaluating MLLMs'  general visual understanding capabilities, which inadequately measures their specific capabilities in human-centric understanding. 
Considering the importance of person as central subjects of world, we construct HERM-Bench to provide comprehensive evaluation  of MLLMs' human-centric understanding capabilities, filling a crucial void in existing benchmarks. 

\section{A Review of Captions on Human-Related Information}
\label{sec3}
Recent studies\cite{zhang2024mm,yu2023capsfusion} have highlighted that the sub-optimal modality alignment, due to lack of high quality image captions, significantly limits the perception and reasoning capabilities of MLLMs. Moreover, simply scaling up monotonous synthetic captions exacerbates model degradation \cite{yu2023capsfusion}. Consistent with these findings, we attribute the unsatisfactory performance of MLLMs in
\textit{human-centric understanding} to the utilization of low-quality human-related descriptions in existing training paradigms \cite{chen2023minigpt,liu2023improved}. In this section, we delve into an analysis of the human-related information present in existing caption datasets.

As one of common visual objects, person appear frequently in mainstream caption datasets \cite{lin2014microsoft, li2022blip}. However, as shown in ~\cref{fig:teaser}, we find their captions suffer from two main shortcomings: (1) failure to provide fine-grained and comprehensive annotations for person; (2) only involving loose individuals appearing in the images. Next, we present quantitative results validating our conjecture.

\begin{figure}[!tbp]
    \centering
    % % 第一个子图
    % \begin{subfigure}[b]{\textwidth}  % 设置宽度为文本宽度的100%
    %     \centering
    %     \includegraphics[width=0.85\textwidth]{Figures/example_coco_caption.pdf}  % 替换成实际图片文件，调整宽度为需要的大小
    %     \caption{}
    %     \label{fig:coco_example}
    % \end{subfigure}
    
    % 下方子图的第一个
    \begin{subfigure}[b]{0.49\textwidth}  % 设置宽度为文本宽度的45%
        \centering
        \includegraphics[width=\textwidth]{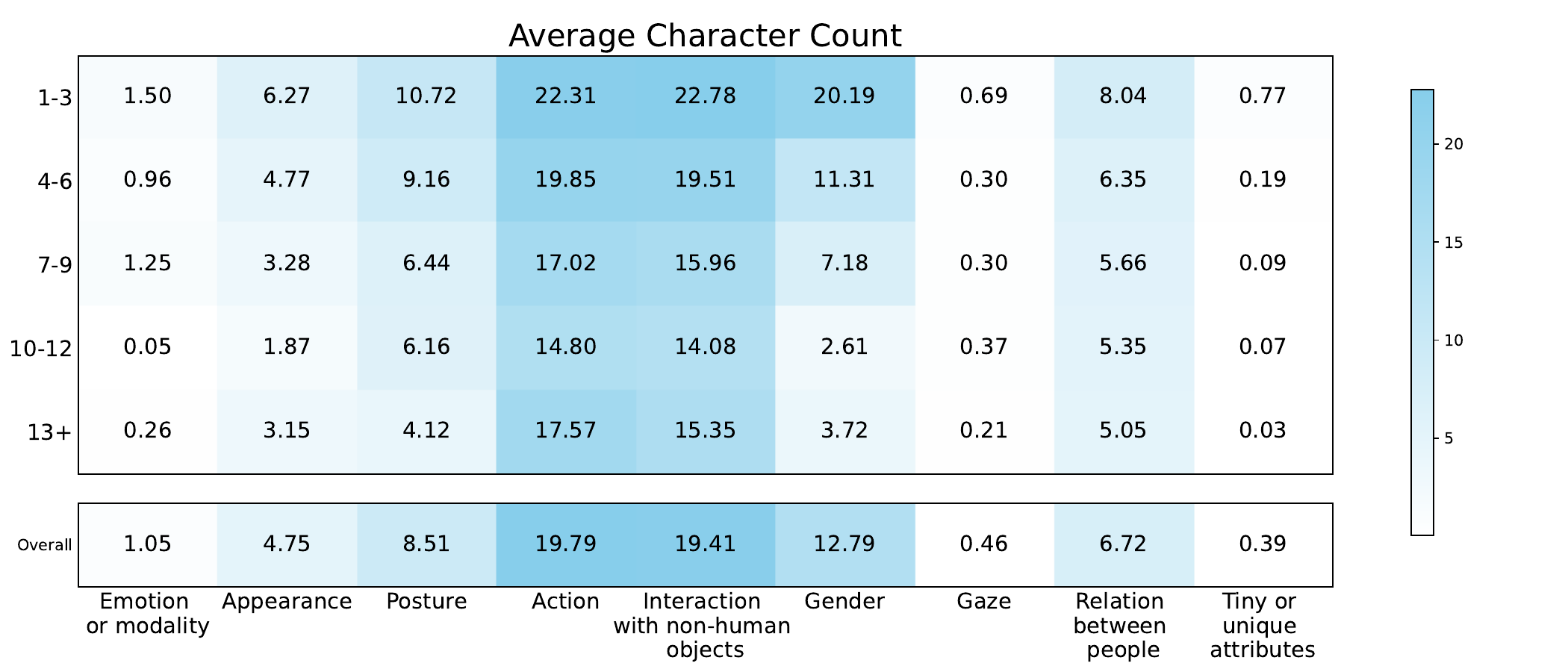}  % 替换成实际图片文件
        \caption{}
        \label{fig:avg_char}
    \end{subfigure}
    %\hfill  % 在两个子图之间添加一些空白
    % 下方子图的第二个
    \begin{subfigure}[b]{0.49\textwidth}  % 设置宽度为文本宽度的45%
        \centering
        \includegraphics[width=\textwidth]{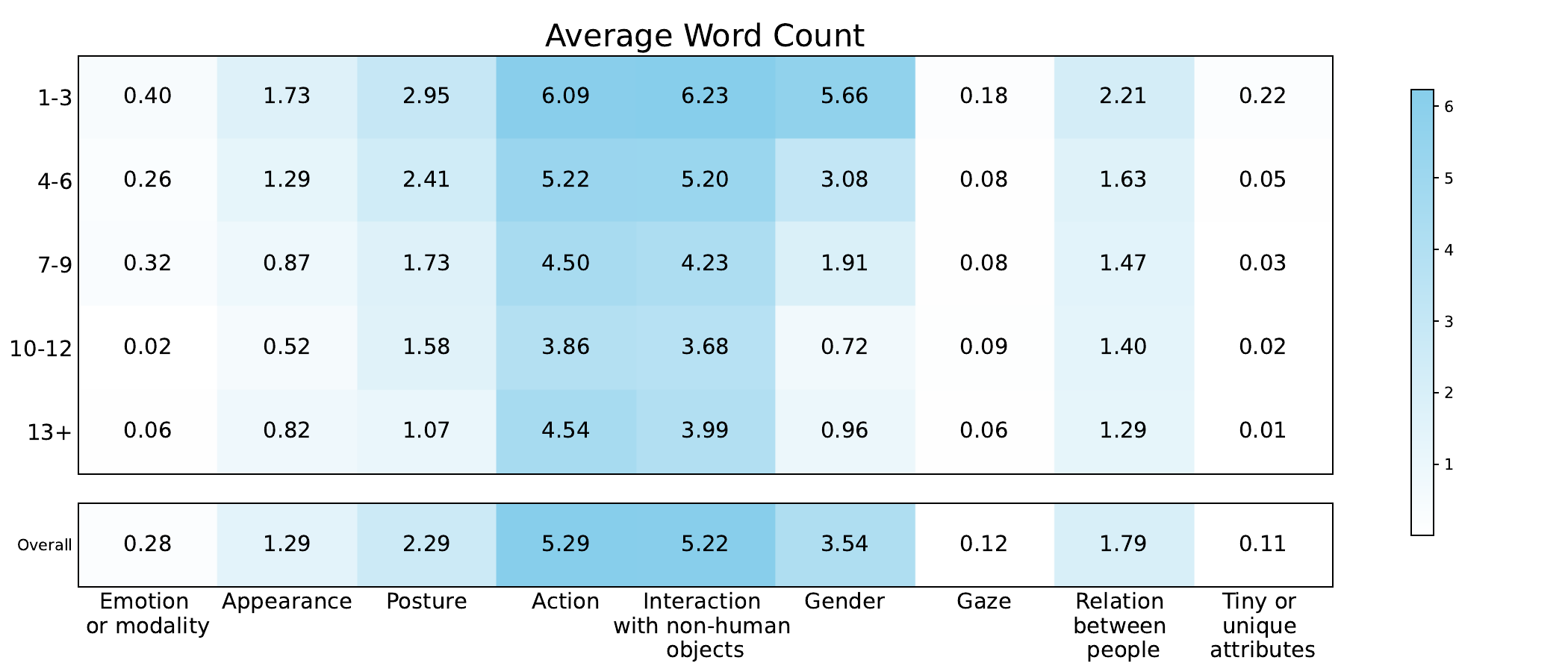}  % 替换成实际图片文件
        \caption{}
        \label{fig:avg_token}
    \end{subfigure}
    \vspace{-5pt}
    \caption{
Human-related information distribution in COCO captions. (a)/(b): heatmaps representing the average number of characters/words to describe various aspects (appearance, action, \etc), grouped by the person number in each image (ranging from 1-3 to 13+). It is observed that descriptions of all sides in COCO are limited to a few words and become increasingly inadequate in scenes with a larger number of people.
}
    \label{fig:review}
    \vspace{-15pt}
\end{figure}

\noindent{\textbf{Quantify the human-related annotation in various aspects.}}
To quantify the quality of human-related information in existing captions, Fig. \ref{fig:review} conducts a detailed statistical analysis of human-related information on a subset of $1000$ samples containing people from COCO Caption \cite{lin2014microsoft}, a dataset widely used for pre-training and generating instruction following data in mainstream MLLMs\cite{liu2023improved,chen2023minigpt,chen2023shikra, you2023ferret}. In details, we firstly employ a hand-crafted list of human visual aspects, \eg, appearance, posture, gender and gaze. Then, we leverage GPT-4 \cite{openai2023gpt} to measure the information content in each aspect, calculating the average length (by character or word) of descriptions per aspect and per individual in the captions. This offers a reference metric for the quantity of information obtained from the captions.

\noindent{\textbf{COCO captions fall short in scope of human-related annotations.}}
As depicted in Fig. \ref{fig:review}, we can observe that: (1) there is a severe \textit{imbalance} in the average description length among different aspects of human-related information. 
These captions predominantly focus on actions of persons and their relations to objects. However, they frequently overlook other essential aspects such as specific appearance, pose of people, and activities among people. (2) The average description length tends to decrease as the number of people in the scene increases, across all aspects. In more complex scenes with more individuals, COCO captions often mention only  a subset of them or use general phrases (like `a group of') to describe person involved in the main activity, while disregarding exceptional individuals. Overall, COCO captions fail to provide comprehensive annotations that capture diverse perspectives on people.

\noindent{\textbf{COCO captions fall short in granularity of human-related annotations.}}
Another notable observation is the lack of fine granularity in COCO captions.  
As shown in Fig. \ref{fig:review}, these captions typically provide brief and coarse-grained descriptions. Fine-grained aspects, such as gaze direction or accessories, are rarely mentioned.   
Even for more common aspects like actions, the average length of descriptions is only about $5$ words. Additionally, these descriptions are limited to coarse terms, \eg, standing and reading, lacking details and supplementary context (such as specific emotions and poses) associated with the actions.

\section{Benchmark MLLMs on Human-Centric Understanding}
\label{sec4}
Given the pivotal role of human-centric learning in the development of MLLMs, it is crucial to establish a benchmark for the quantitative evaluation of MLLMs on comprehensive human-centric tasks. Moreover, as revealed in Sec.~\ref{sec3}, the human-related information in current MLLM training data suffers from limited scope and granularity. This raises an urge need to assess whether the human-centric capabilities of current MLLMs are impeded by the drawbacks in training data.

In this work, we propose the first MLLM benchmark, named HERM-Bench, specialized on human-centric domain. The contributions of HERM-Bench are two-fold: (1) It provides a comprehensive quantitative evaluation spanning $8$ human-centric dimensions, covering both basic perception and complex understanding capabilities. (2) By evaluating MLLMs on HERM-Bench, we have confirmed that current MLLMs fall short in full-scope and fine-grained human-centric knowledge, which reinforces our proposal of improving the quality of human-related annotations in training data. In this section, we provide the task taxonomy and construction pipeline of HERM-Bench, and conduct preliminary evaluations on existing MLLMs.

\begin{figure}[tb]
    \includegraphics[width=0.95\linewidth]{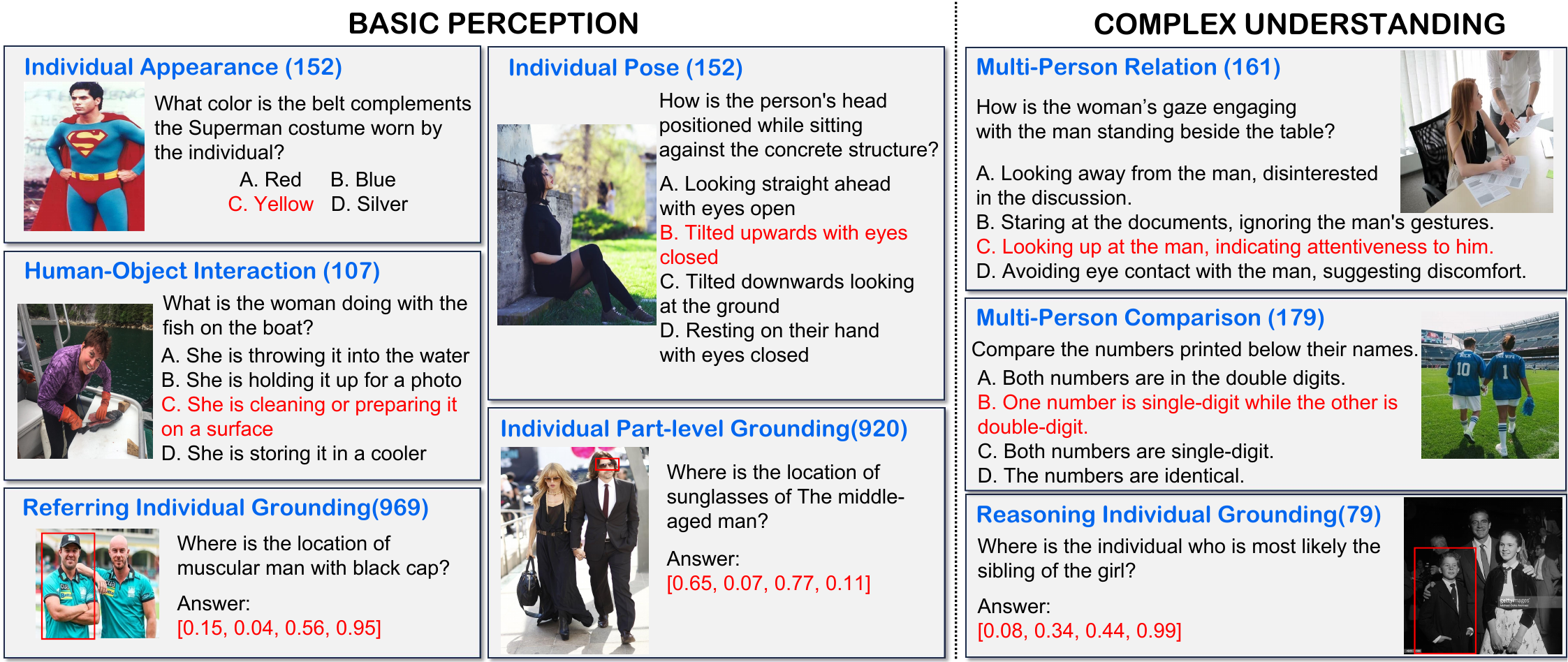}
      \vspace{-5pt}
    \caption{Taxonomy and examples of HERM-Bench. HERM-Bench includes 8 evaluation dimensions on basic perception and complex understanding fields. The number in bracket denotes question number of each evaluation dimension.}
    \label{fig:test-benchmark-taxonomy}
    \vspace{-15pt}
\end{figure}

\subsection{Task Taxonomy}
\label{sec:Task_Taxonomy}
\textbf{Task Dimensions.} \ 
To comprehensively assess human-centric capabilities of MLLMs, HERM-Bench incorporates $8$ evaluation dimensions encompassing both basic perception and complex understanding, as shown in Fig.~\ref{fig:test-benchmark-taxonomy}.

\noindent
1. \emph{\textbf{Basic Perception.}} It refers to directly acquiring visual information of person in the image. We consider $5$ dimensions covering appearance, pose, human-object interaction and grounding of single individual:
\begin{itemize}
    \item[$\bullet$] Individual Appearance (\textbf{IA}): \ Recognize the visual appearance of a specified individual in the image, such as hairstyle and outfits.
    \item[$\bullet$]  Individual Pose (\textbf{IP}): \ Identify the body posture of a specified person, such as body orientation and position of body parts. 
    \item[$\bullet$]  Human-object Interaction (\textbf{HOI}): \ Identify the interactions between a specified person and other non-human objects within the image.  
    \item[$\bullet$]  Referring Individual Grounding (\textbf{REF}): \ Locate a specific person based on explicit attributes such as appearance and pose.
    \item[$\bullet$]  Individual Part-level Grounding (\textbf{IPG}): \  Locate a certain element of a specific person, such as clothes, body part and accessories.
\end{itemize}
2. \emph{\textbf{Complex Understanding.}} It refers to analyzing and reasoning based on the basic perception information of humans. We consider $3$ dimensions focusing on the relation, comparison and reasoning among multiple individuals:
\begin{itemize}
    \item[$\bullet$] Multi-Person Relation (\textbf{MPR}): \ Understand various relations among multiple individuals within the image, such as interactions and social relations.
    \item[$\bullet$] Multi-Person Comparison (\textbf{MPC}): \ Analyze the commonness or difference among individuals, such as commonness/differences in clothes and identities. 
    \item[$\bullet$] Reasoning Individual Grounding (\textbf{REA}): \ Locate the person based on implicit references, such as his/her relationship to other person and  role in a group. This requires the model to reason the reference to identify the target. 
\end{itemize}

\noindent{\textbf{Output Formats.}} \
To take both natural language description and spatial grounding abilities into account, HERM-Bench consists of two task formats: (1) Multiple-Choice questions. This format assesses the model's natural language proficiency by constructing questions with multiple choices in natural language form. (2) Grounding questions. These questions aim at evaluating the model's capability to identify and specify the spatial positions of individuals within an image, with responses expected in the form of bounding box coordinates.

\begin{figure}[!tbp]
    \centering
    \includegraphics[width=0.8\linewidth]{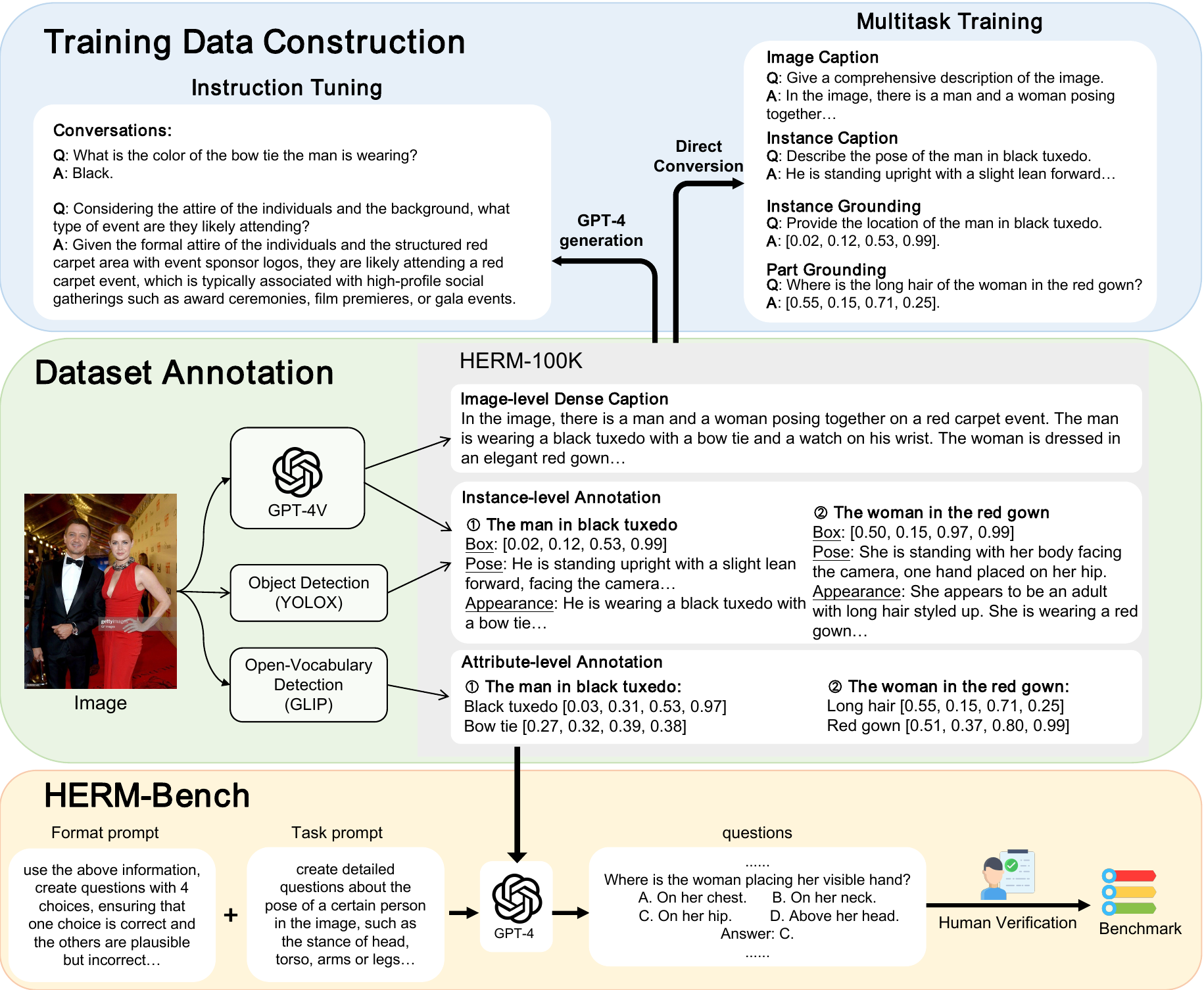}
    \caption{The overall pipeline of constructing HERM-100K, HERM-Bench and training data. First, we derive HERM-100K using powerful off-the-shelf foundation models. Then, using visual annotations in HERM-100K, we create multitask training and instruction tuning data, as well as prompting GPT-4 to build HERM-Bench.}
    \label{fig:pipeline}
    \vspace{-17pt}
\end{figure}

\subsection{Benchmark Construction}
\label{sec:benchmark construction}
As~\cref{fig:pipeline} shows, we employ a GPT-assisted pipeline for benchmark construction, involving question-answer generating and verification.  
We firstly utilize GPT-4V~\cite{OpenAI2023GPT4v} to extract comprehensive visual information, and subsequently use GPT-4~\cite{openai2023gpt} to generate question-answer pairs for each evaluation dimension. To ensure the quality and reliability of these pairs, we filter out low-quality question-answer pairs. This pipeline ensures that HERM-Bench is not only comprehensive but also precise and reliable in measuring the intended competencies.

\noindent
\textbf{Visual Information Collection.} \ 
To collect diverse visual information for question generation, we use a variety of image sources, including CC~\cite{sharma2018conceptual}, SBU~\cite{saleh2015large} and LAION~\cite{schuhmann2021laion} datasets. As shown in~\cref{fig:pipeline}, to ensure the richness of visual information, we meticulously curated image annotations encompassing human information at full scopes and levels (details introduced in~\cref{sec:training_annotation}). These annotations, with comprehensive and fine-grained description of human information, can enhance the breadth and depth of posed questions.

\noindent
\textbf{Question Generation.} \ 
Based on the extracted visual information, we prompt GPT-4 to generate questions for each evaluation dimension, as shown in~\cref{fig:pipeline}. We employ different \emph{format prompts} for multi-choice questions and grounding questions. Additionally, we develop specific \emph{task prompts} for each task dimension. When generating questions for each task dimension, we combine the corresponding \textit{format prompt} with the \textit{task prompt}, creating a comprehensive prompt for GPT-4. The detailed prompts are provided in Appendix.

\noindent
\textbf{Human Verification.} \
To ensure the quality of HERM-Bench, we employ human annotators to verify the generated question-answer pairs. Annotators are asked to answer each question. Any question that cannot be answered based on the visual concept, can be answered without resorting to input image, or deviates from corresponding task dimension is discarded by the annotators. This process yields a clean, high-quality benchmark for evaluation, comprising a total of $2,748$ questions. See~\cref{fig:test-benchmark-taxonomy} for question number on each evaluation dimension.

\noindent
\textbf{Evaluation Strategy.} \
We adopt different evaluation strategies for multi-choice and grounding questions. 
For multiple-choice questions, following~\cite{ji2023large}, we provide the question, options, and the answer generated by the MLLMs to GPT-4, prompting it to select the option closest to MLLM's output. For grounding questions, we compute accuracy based on the IoU between predicted box and ground truth. Following previous works~\cite{chen2023minigpt, chen2023shikra}, the IoU threshold is set to $0.5$.

\subsection{Preliminary Evaluations of Existing MLLMs on HERM-Bench}
\label{sec:preliminary evaluation}
To preliminarily explore the capabilites of existing MLLMs on human-centric tasks, we use questions from HERM-Bench to inquire existing MLLMs. Fig.~\ref{fig:experiment_examples} shows some examples. We observe that: (1) For questions related to overlooked human aspects in current training data, existing MLLMs yield unsatisfactory answers. For example, existing training data overlook \textit{posture} and \textit{relations between people} (\cref{fig:review}), and we note that existing MLLMs perform  poorly on questions regarding \emph{body pose} (\cref{fig:experiment_examples}(b)) and \emph{multi-person relation} (\cref{fig:experiment_examples}(d)). (2) For questions with fine-grained human-related aspects that are absent from current training data, existing MLLMs fail to accurately solve them. For example, when asked about fine-grained appearance details, existing MLLMs give incorrect answers (\cref{fig:experiment_examples}(a)). 
In conclusion, the evaluation results suggest that sub-optimal training data directly limits the human-centric capabilities of MLLMs.

\section{Enhance Human-centric Understanding from Better Captions}
As discussed in Sec. \ref{sec3} and Sec. \ref{sec4}, in current MLLMs training data, the inferior quality of human-related captions poses a severe limitation to the human-centric knowledge of existing MLLMs. To solve this problem, we propose HERM-100K, highlighted by its improved scope and granularity of human-related annotations. In this section, we introduce the pipeline of building HERM-100K, and the construction of training data from HERM-100K.

\subsection{Construction Pipeline of HERM-100K} 
\label{sec:training_annotation}
To improve human-related annotations, we create a new human-centric dataset, named HERM-100K, by prompting GPT-4V\cite{OpenAI2023GPT4v}, a powerful MLLM, to generate diverse annotations from three levels of visual contents. As in Fig. \ref{fig:pipeline}, the annotations consist of (1) image-level dense captions that provide a thorough understanding of visual scene. (2) instance-level captions that capture multiple dimensions of the individual. (3) attribute-level annotations that highlight specific body-parts or rare attributes. Notably, for instance-level and attribute-level, each annotation is linked with a region in the image. Next, we delve into the details of the three levels of annotations.

\begin{itemize}

\item[$\bullet$] \textbf{Level-1: Dense caption for image-level understanding} In level-1, we prompt GPT-4V to generate a comprehensive description of the scene, highlighting people and other objects. We also encourage the model to depict interaction among people or objects, and indicate specific events and locations if they can be confidently identified, so as to obtain a faithful caption of the visual content, as well as open-world knowledge implied in the image. 

\item[$\bullet$] \textbf{Level-2: Multi-perspective instance-level descriptions}
In level-2, we equip each person in the image with captions of diverse perspectives, including \textit{appearance}, \textit{pose}, \textit{modality} and \textit{spatial or interactive relations} with other objects. In implementation, we present GPT-4V with the whole image, followed by cropped patches of bounding boxes for each person, prompting GPT-4V to annotate each person from above perspectives (for images without human bounding box, we employ a light-weight detection model YOLOX~\cite{ge2021yolox} to detect persons and use its predictions as pseudo annotation). In this way, the model could focus on generating instance-level annotation while avoiding ambiguity and the lack of context in cropped patches.

\item[$\bullet$] \textbf{Level-3: Attribute-level phrases within a person} In level-3, we focus on adding attribute-level annotations, such as body-parts, clothing and accessory. These attributes are partly sourced from the original annotations of images in human-parsing tasks\cite{tang2023humanbench}, which provide masks and labels for body-parts and clothing. Additionally, we use GPT-4 to parse attributes from level-2 description, to identify more specific clothing and rare attributes. Pseudo-region annotations are also provided by an open-vocabulary detection model, GLIP~\cite{li2022grounded}. Finally, these attributes are linked to reference expressions of each instance (details see Sec \ref{sec:training_data}), creating complete phrases.

\end{itemize}

 \noindent{\textbf{Data Source.}} To ensure the diversity of data, the images come from four sources: COCO images\cite{lin2014microsoft} containing people; human pose estimation dataset AIC\cite{wu2017ai} with part annotations; human parsing dataset CIHP\cite{gong2018instance} and web image-text dataset CCS-LAION\cite{li2022blip}. To reach a trade-off between diversity and accuracy, we apply a heuristic rule to filter images, \eg, low-resolution images. The detailed image statistics and heuristic rule are provided in Appendix.

\noindent{\textbf{Prompt Design.}} \ To prompt GPT-4V to generate instance-level captions, we first present GPT-4V the original image for generating level-1 caption. Then, we input cropped patches of each instance to generate instance-level annotation, with  the context of the first-round visible. The prompt templates and discussion on other region prompt strategies (like SoM\cite{yang2023set}) are provided in Appendix.

After filtration, we obtain HERM-100K with 10,609 image-level captions, 21,489 instance-level captions and 97,320 attribute-level annotations. For image-level/instance-level captions, the average word count is 120.6/81.8, largely surpassing COCO caption with 12.0 words on average. For attribute-level annotation, each instance is equipped with 3.53 attributes on average per person, drawn from 6017 unique attribute phrases, surpassing existing human parsing or attribute recognition datasets with only dozens of attribute classes.

\subsection{Construct Training Data from HERM-100K}
\label{sec:training_data}
In the common training scheme of MLLMs\cite{zhu2023minigpt, zhao2023svit, liu2024visual}, image captions play two roles, \ie multi-modal alignment during pre-training\cite{li2022blip, lin2014microsoft, chen2023sharegpt4v} and creating instruction-following datasets for supervised fine-tuning\cite{liu2024visual,zhao2023svit, chen2023shikra, wang2023see}. Therefore, we formulate the annotations from HERM-100K (Sec \ref{sec:training_annotation}) into a variety of question-answer tasks, and integrate them into both pre-training and instruction tuning processes of MLLMs.

\noindent{\textbf{Multitask Training Data.}} To refer to specific instance in questions and answers, beyond the bounding boxes, we extracted \textit{diverse referential expressions} from instance-level annotations by LLM. Utilizing these reference expressions, the multi-level annotations can be seamlessly formulated into \emph{image-level caption} and \emph{instance-level caption} tasks. This formulation is achieved by posing questions about a specific region or whole image, and answering with the annotations of corresponding level,  as illustrated in Fig.~\ref{fig:pipeline}. Moreover, region annotations can be readily converted into \emph{instance-level grounding} and \emph{part-level grounding} tasks, by providing the model with reference expressions or phrases, and asking it to output the bounding box of the region. 

\noindent{\textbf{Instruction Tuning Data.}} \
Previous works like LLaVA\cite{liu2024visual,liu2023improved} commonly utilize COCO annotations to prompt ChatGPT\cite{openai2023gpt} to generate conversation data for instruction tuning. Following these works, we generate diverse conversations and complex-reasoning questions via GPT-4\cite{openai2023gpt}, albeit from \textit{our enriched annotations}, as shown in~\cref{fig:pipeline}. To specify detailed and spatial-related information, we also encourage including bounding box of instance or attributes within questions and answers\cite{you2023ferret}. Owing to the comprehensive human-centric annotations, the instruction tuning data is deeply related to human visual contents, as shown in Fig.~\ref{fig:pipeline}. This is critical for MLLMs to understand instructions and use human visual information to perform understanding and reasoning on the image\cite{liu2023improved,you2023ferret}.

In total, we create 320K multitask training data from 6,982 images and 29K instruction tuning pairs from 3,627 images. The average lengths of questions and answers from instruction tuning data are 14.5 and 27.5 words, respectively. More details and statistics are provided in the Appendix.

\section{Experiments}
\subsection{Experiments setup}
\noindent{\textbf{Implementation details.}} \ 
Our model builds upon the established architecture and implementation strategy of MiniGPT-v2~\cite{chen2023minigpt}, which consists of a CLIP encoder, a linear projector, and a Llama-2-7B language model. Specifically, we initialize our model from the stage 2 checkpoint of MiniGPT-v2\footnote{\href{https://drive.google.com/file/d/1Vi_E7ZtZXRAQcyz4f8E6LtLh2UXABCmu/view?usp=sharing}{Checkpoint of MiniGPT-v2 (after stage-2)}} and use 448$\times$448 image resolution via the strategy of concatenating every 4 visual tokens in both training and evaluating phases following \cite{chen2023minigpt}. Subsequently,  we conduct training in two stages. In the multitask training stage, we train the model on our constructed captioning and grounding data (Sec. \ref{sec:training_data}), mixed with other datasets originally used by MiniGPT-v2 including a range of VQA\cite{hudson2019gqa,marino2019ok,schwenk2022okvqa,goyal2017making,rogez2019lcr}, grounding\cite{kazemzadeh2014referitgame,mao2016generation,yu2016modeling} and caption\cite{lin2014microsoft}. In the instruction tuning stage, we finetune the model on our generated instruction-following data (\cref{sec:training_data}). mixed with other instruction-following datasets~\cite{honovich2022unnatural, liu2024visual}. In both stages, we only tune the linear projector and the large language model using LoRA~\cite{hu2021lora}. The illustration of these datasets, data-mixing strategy and detailed training configurations are provided in Appendix.

\begin{table}[t]
    \centering
        \caption{Performance comparison on HERM-Bench. We report accuracy for multiple-choice questions and Acc@0.5 for grounding questions. `-' denotes that MLLMs are unable to evaluate on grounding tasks.}
          \vspace{-5pt}
        \resizebox{0.8\textwidth}{!}{
        \begin{tabular}{l|p{1.0cm}<{\centering}p{1.0cm}<{\centering}p{1.0cm}<{\centering}p{1.0cm}<{\centering}p{1.0cm}<{\centering}|p{1.5cm}<{\centering}p{1.5cm}<{\centering}p{1.5cm}<{\centering}}
        \toprule
        \multirow{2}*{Method} & \multicolumn{5}{c|}{\textbf{Basic Perception}} & \multicolumn{3}{c}{\textbf{Complex Understanding}} \\
        \cline{2-9} & IA & IP & HOI & REF & IPG & MPR & MPC & REA   \\ \midrule
        LLaVA~\cite{liu2024visual}       & 59.2 & 56.5 & 71.0 & - & - & 63.9 & 56.4 & - \\
        LLaVA-1.5~\cite{liu2023improved}    & \underline{75.7} & 61.1 & 72.8 & - & - & 67.1 & 59.2 & - \\
        BLIP-2~\cite{li2023blip}       & 60.5 & 50.6 & 65.4 & - & - & 61.5 & 60.9 & - \\
        InstructBLIP~\cite{dai2023instructblip} & 66.4 & 57.2 & 63.5 & - & - & 66.5 & 59.8 & - \\
        Qwen-VL-Chat~\cite{bai2023qwen} & 61.8 & 62.5 & \underline{76.6} & 17.7 & 14.6 & 69.6 & 54.7 & 27.8 \\
        Shikra~\cite{chen2023shikra}       & 71.0 & \underline{65.1} & 73.8 & 56.1 & 10.0 & 65.2 & \underline{61.4} & 51.5 \\
        InternLM~\cite{team2023internlm}     & 70.4 & 55.9 & 69.2 & - & - & 65.2 & 57.6 & - \\
        Kosmos-2~\cite{peng2023kosmos}      & 57.2 & 61.8 & 76.5 & 40.9 & 12.0 & \underline{69.5} & 57.5 & 29.1 \\
        Ferret~\cite{you2023ferret}         & 73.6 & 62.5 & 73.8 & \underline{58.7} & 15.1 & 64.5 & 53.1 & \underline{53.1} \\
        % NextChat~\cite{zhang2023next}       & & & & 59.1 & 15.8 & & & 53.1 \\
        OFA-H~\cite{wang2022ofa}            & 73.3 & 57.2 & 72.9 & 54.8 & \underline{41.0} & 60.2 & 53.1 & 24.1 \\
        MiniGPT-v2~\cite{chen2023minigpt}   & 70.4 & 63.8 & 68.2 & 57.2 & 29.7 & 59.0 & 55.3 & 44.3 \\ \midrule
        HERM-7B (ours)            & \textbf{82.2} & \textbf{72.4} & \textbf{82.2} & \textbf{59.8} & \textbf{42.6} & \textbf{72.0} & \textbf{68.7} & \textbf{54.4} \\
    
        \bottomrule
 	\end{tabular}}
	\label{tab:Benchmark_expt_results}
        \vspace{-17pt}
\end{table}

\noindent{\textbf{Evaluation Setup.}} \ 
Besides our constructed HERM-Bench, we also evaluate HERM-7B on common VQA and Reference Expression Comprehension (REC) benchmarks. For VQA benchmarks, we choose OKVQA\cite{marino2019ok} and GQA\cite{hudson2019gqa}. For REC benchmarks, we evaluate on RefCOCO~\cite{kazemzadeh2014referitgame}, RefCOCO+~\cite{yu2016modeling}, and RefCOCOg~\cite{mao2016generation}. 
For evaluation on these datasets, we opted for original test configurations as MiniGPT-v2, reporting top-1 accuracy and Acc@0.5 respectively. We encompassed a range of previous works for a thorough comparison, including (1) MLLMs that can generate open-ended text, such as LLaVA-7B\cite{liu2024visual}, LLaVA-1.5-7B\cite{liu2023improved}, InstructBLIP-Vicuna7B\cite{dai2023instructblip}. (2) MLLMs that can refer and ground locations beyond open-ended text, such as Shikra-7B\cite{chen2023shikra}, Ferret-7B\cite{you2023ferret}.
The detailed illustrations for these datasets and models are provided in the Appendix.

\subsection{Quantitative Results}
\noindent
\textbf{Comparisons on HERM-Bench.} \
We firstly conduct a comprehensive evaluation on HERM-Bench, quantitatively comparing the performance of our proposed HERM-7B with existing state-of-the-art MLLMs. As~\cref{tab:Benchmark_expt_results} shows, HERM-7B attains the most superior performance across all 8 task dimensions, demonstrating exceptional advantage in both basic perception and complex understanding abilities. Specifically,
for \textbf{basic perception} tasks, our model achieves an average gain of $9.98\%$ over baseline MiniGPT-v2~\cite{chen2023minigpt}. In details, HERM-7B encompasses superiority in recognizing individual traits (IA, IP), identifying human interaction with objects (HOI), and localizing humans and their part-level attributes (REF, IPG). This verifies the robust advantage of our model in various human-centric perception abilities. 
For \textbf{complex understanding} tasks, our model achieves the best accuracy on all 3 task dimensions (MPR, MPC, REA), with an average  gain of $12.2\%$.
The results validate the substantial potential of our model in understanding and reasoning about complex human-centric scenes.

Further, we observe existing MLLMs tend to  excel in only one or two tasks (\eg, Qwen-VL-Chat~\cite{bai2023qwen} on HOI), while performing relatively poorly on other tasks (\eg, Qwen-VL-Chat on IA, IP). In contrast, our model maintains a consistent advantage across all tasks, further consolidating its capability to encompass comprehensive human-centric knowledge.

\begin{table}[t]
    \centering
    \scriptsize
    \begin{minipage}[t]{0.35\textwidth}
        \centering
        \caption{Performance comparisons on VQA benchmarks. The results of baselines are from \cite{chen2023minigpt}.}
        % \setlength\tabcolsep{2pt}
        % \small
        %\renewcommand{\arraystretch}{1.25}
        \resizebox{0.9\textwidth}{!}{
        \begin{tabular}{l|c|c}
            \toprule
            Method           & OKVQA & GQA  \\ \midrule
            % Flamingo-9B~\cite{alayrac2022flamingo}      & 44.7  & -    \\
            BLIP-2~\cite{li2023blip}       & 45.9  & 41.0 \\
            InstructBLIP~\cite{dai2023instructblip} & -     & 49.5 \\
            MiniGPT-4~\cite{zhu2023minigpt}    & 37.5  & 30.8 \\
            LLaVA~\cite{liu2024visual}      & 54.4  & 41.3 \\
            Shikra~\cite{chen2023shikra}       & 47.2  & -    \\
            % mPLUG-Owl2~\cite{ye2023mplug}       & 57.7  & 56.1 \\
            MiniGPT-v2~\cite{chen2023minigpt}    & \textbf{58.0} & \textbf{59.5} \\  \midrule
             HERM-7B (ours)             & 55.4  & 58.4 \\
            \bottomrule
        \end{tabular}
        }
        \label{tab:VQA_expt_results}
    \end{minipage}\hfill
    \begin{minipage}[t]{0.63\textwidth}
        \centering
        %\vspace{-15pt}
        \caption{Performance comparisons on REC benchmarks. The results of baselines are from \cite{chen2023minigpt}. Other than the enhanced ability in grounding human-related items, our model is  also capable of grounding regions from a common reference expression.}
        % \setlength\tabcolsep{2pt}
        % \small
        %\renewcommand{\arraystretch}{1.25}
        \resizebox{0.995\textwidth}{!}{
        \begin{tabular}{l|ccc|ccc|cc|c}
            \toprule
            Method        & \multicolumn{3}{c|}{RefCOCO} & \multicolumn{3}{c|}{RefCOCO+} & \multicolumn{2}{c|}{RefCOCOg} & Avg\\ 
                          & val   & test-A   & test-B    & val   & test-A   & test-B     & val    &  test                &    \\ \midrule
            OFA-L~\cite{wang2022ofa}         & 79.96 & 83.67 & 76.39 & 68.29 & 76.00 & 61.75 & 67.57 & 67.58 & 72.65 \\ 
            Shikra~\cite{chen2023shikra}     & 87.01 & 90.61 & 80.24 & \textbf{81.60} & \textbf{87.36} & 72.12 & 82.27 & 82.19 & 82.93 \\ 
            MiniGPT-v2~\cite{chen2023minigpt} & \textbf{87.23} & \textbf{91.21} & 83.59 & 78.79 & 85.14 & \textbf{72.94} & \textbf{83.35} & \textbf{84.20} & \textbf{83.31} \\ \midrule
             HERM-7B (ours)          & 86.56 & 89.43 & \textbf{83.65} & 78.31 & 82.90 & 72.18 & 82.01 & 82.73 & 82.22 \\
            
            \bottomrule
        \end{tabular}
        }
        \label{tab:REC_expt_results}
    \end{minipage}
    \vspace{-17pt}
\end{table}

%\subsection{Comparisons on General Vision-Language Tasks}
\noindent
\textbf{Comparisons on general vision-language tasks.} \
While training HERM-7B to acquire human-centric knowledge, it is also critical for HERM-7B to retain \textit{general knowledge}. To assess the capability of our HERM-7B in general knowledge domain, we evaluate its performance on general VQA and REC tasks, aligning with baseline MiniGPT-v2~\cite{chen2023minigpt}. 
%\cref{tab:VQA_expt_results} and \cref{tab:REC_expt_results}   present the comparisons on VQA and REC tasks. 
As shown in \cref{tab:VQA_expt_results} and \cref{tab:REC_expt_results}, HERM-7B \textit{slightly} lags behind state-of-the-art MLLMs on general vision-language tasks.  The results confirm that besides excelling in human-centric understanding,  HERM-7B still maintains a strong capability in the general knowledge domain.

\begin{figure}[t]
    \centering
    \includegraphics[width=0.9\textwidth]{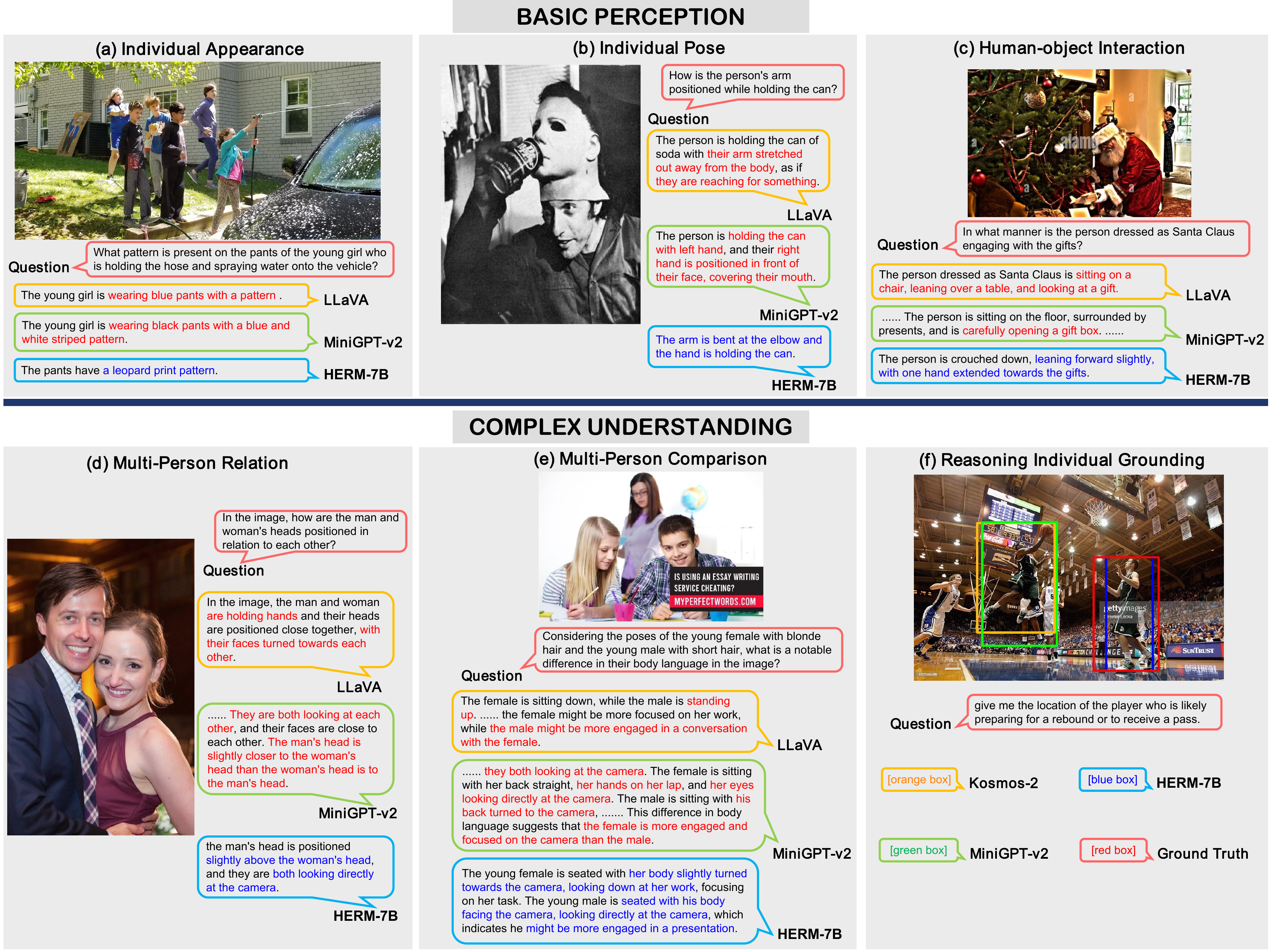}  
    \caption{Evaluation examples on HERM-Bench. We compare outputs of LLaVA, MiniGPT-v2 and HERM-7B. We mark error parts in {\color{red} red}, while correct parts in {\color{blue} blue}.}
    \label{fig:experiment_examples}
    \vspace{-10pt}
\end{figure}

\subsection{Qualitative Results}
Fig.~\ref{fig:experiment_examples} provides qualitative comparisons of HERM-7B against existing MLLMs (LLaVA and MiniGPT-v2). We can observe that, firstly, \textbf{HERM-7B exhibits satisfying basic perception abilities on \textbf{fine-grained} human-centric information from full perspectives.} In details, {\color{red} (a)} shows HERM-7B excels at discerning intricate appearance details (\textit{the leopard print pattern}), while other MLLMs misinterpret the \textit{color} and \textit{style} of the pattern; {\color{red} (b)} shows that HERM-7B can accurately recognize delicate arm pose (\textit{bent at the elbow}), while other models fail to perceive pose caused by misconception of mask; In {\color{red} (c)}, HERM-7B correctly identifies engagement between santa and gift (\textit{extended towards}), but other models simply misinterpret the nature of minute physical interaction (\textit{looking/opening the gift}). Secondly, \textbf{HERM-7B possesses robust understanding of complex human-related scenarios.} In details, {\color{red} (d)} shows HERM-7B accurately understands the relative head position of the two people. But other MLLMs \textbf{\textit{wrongly perceive}} both people's head postures, thus misunderstanding their relative head position; In {\color{red} (e)}, HERM-7B precisely analyzes the difference in body language between two people, based on correctly perceiving their body positions and gaze directions. However, other models fail at correctly perceiving these body status, leading to wrong understanding of human differences. In {\color{red} (f)}, HERM-7B accurately reasons the person location based on his role in basketball game (\textit{to receive a pass}), while other models locate the person shooting the ball,  possibly caused by wrong perception of complex body pose in basketball game.

\begin{table}[t]
	\centering
	\caption{Ablation study on the effect of data quality in different training stages. Incorporating data derived from HERM-100K into both training stages enhances performance, with the combination of both yielding even superior improvement.}
        % This indicates the cruciality of data quality and advantages of HERM-100K.
        \setlength\tabcolsep{2pt}
        \small 
        \resizebox{0.9\textwidth}{!}{
        \begin{tabular}{cp{3cm}<{\centering}|p{1cm}<{\centering}p{1cm}<{\centering}p{1cm}<{\centering}p{1cm}<{\centering}p{1cm}<{\centering}p{1cm}<{\centering}p{1cm}<{\centering}p{1cm}<{\centering}|p{1cm}<{\centering}}
        \toprule
        Pre-training & Instruction-tuning & IA & IP & HOI & REF & IPG & MPR & MPC &  REA & Avg. \\ \midrule
        \XSolidBrush & \XSolidBrush & 67.8 & 61.2 & 59.8 & 54.1 & 28.0 & 62.1 & 59.8 &  39.2 & 54.0 \\
        \XSolidBrush & \Checkmark & 79.6 & 67.1 & 80.3 & 57.7 & 36.4 & 72.0 & 65.4 &  45.5 & 63.0 \\
        \Checkmark & \XSolidBrush & 72.4 & 67.6 & 80.3 & 59.3 & 41.5 & 71.4 & 64.8 &  51.0 & 63.5 \\
        \midrule
        \Checkmark & \Checkmark & \textbf{82.2} & \textbf{72.4} & \textbf{82.2} & \textbf{59.8} & \textbf{42.6} & \textbf{72.0} & \textbf{68.7} &  \textbf{54.4} & \textbf{66.8} \\
        \bottomrule
 	\end{tabular}}
	\label{tab:ablation on training stages}
        % \vspace{-15pt}
\end{table}

\begin{table}[t]
	\centering
	\caption{Ablation study on the impact of various types of annotations in HERB-100K. Instance captioning, instance grounding, and part-level grounding correspond to questions derived from instance-level descriptions, extracted reference expressions, and attribute phrases respectively. }
        \setlength\tabcolsep{2pt}
        \small
        \resizebox{0.9\textwidth}{!}{
        \begin{tabular}{l|p{1cm}<{\centering}p{1cm}<{\centering}p{1cm}<{\centering}p{1cm}<{\centering}p{1cm}<{\centering}p{1cm}<{\centering}p{1cm}<{\centering}p{1cm}<{\centering}|p{1cm}<{\centering}}
        \toprule
        Model & IA & IP & HOI & REF & IPG  & MPR & MPC &  REA & Avg.   \\ \midrule 
        w/o instance captioning  & 75.9 & 65.8 & 68.2 & 58.7 & 41.8 & 64.5 & 64.8 &  49.4 & 61.1 \\
        w/o instance grounding   & 78.3 & 67.8 & 75.7 &  56.7 & 42.4 & 70.0 & 64.2 &  49.4 & 63.1 \\
        w/o part-level grounding & 80.2 & 72.3 & 80.3 &  59.2 & 28.9 & 71.4 & 66.5 &  53.1 & 64.0 \\
        \midrule
        HERM-7B (Ours) & \textbf{82.2} & \textbf{72.4} & \textbf{82.2} & \textbf{59.8} & \textbf{42.6} & \textbf{72.0} & \textbf{68.7} &  \textbf{54.4} & \textbf{66.8} \\
        \bottomrule
 	\end{tabular}}
	\label{tab:ablation on training data}
        \vspace{-15pt}
\end{table}

\subsection{Ablation Study}
\noindent{\textbf{Impact of data quality.}} \
~\cref{tab:ablation on training stages} presents an ablation study to assess the influence of multitask training data and instruction tuning data constructed from HERM-100K.  
Involving each of them into training both leads to significant improvement on HERM-bench, while the combining them yields even better performance. This indicates that the effect of data quality lies both in pre-training and instruction-tuning, and validates the high quality of HERM-100K.

\noindent{\textbf{Effectiveness of multi-level annotations.}} \
~\cref{tab:ablation on training data} presents an ablation study on the impact of various types of annotations from HERM-100K. Excluding questions derived from each type of annotations would lead to a significant drop on the performance on HERM-bench, implying  the necessity of multi-level annotations within HERM-100K.

\section{Conclusion} 
In this study, we focus on exploring MLLMs' capability in human-centric visual understanding. To thoroughly assess this capability, we introduce HERM-bench, the first human-centric MLLM benchmark, extensively covering various human-related task dimensions.  Through benchmark evaluation and analysis, we identify a  significant deficiency in existing MLLMs in term of human-centric knowledge, which can be attributed to low-quality human-related annotations. 
As a solution, we propose HERM-100K including multi-level comprehensive human-related annotations. By integrating HERM-100K into MLLMs' training, we 
observe a substantial performance gain in human-centric tasks. 
Our work sheds light on the untapped potential of MLLMs in human-centric tasks and provides a  foundation for future research in human-related  video understanding and AIGC \cite{tevet2022human}. 

\setcounter{table}{0}
\setcounter{section}{0}
\setcounter{figure}{0}
\renewcommand{\thefigure}{S\arabic{figure}}
\renewcommand{\thetable}{S\arabic{table}}
\renewcommand\thesection{\Alph{section}}

\section*{Appendix}
\appendix

\section{Related Work}
\noindent{\textbf{Human-Centric Foundation Models.}} \
Human-centric foundation models aim to develop universal models capable of addressing various traditional human-centric visual tasks, \eg, person re-identification~\cite{Zhu_Wu_Huang_Zheng_2018, hou2021feature}, pose estimation~\cite{rogez2019lcr} and human parsing~\cite{liang2018look}. 
For instance, HCMoCo~\cite{hong2022versatile}  attempts to obtain universal representations by harnessing multi-modal human data. 
PATH~\cite{tang2023humanbench} trains a general backbone with specific projector for each human-centric task. 
\cite{ci2023unihcp, wang2023hulk}  focus on simultaneously processing traditional human-centric tasks within a unified model. However, these works  are constrained to predefined perception tasks and lack the flexibility to address free-form questioning and various visual understanding tasks. In contrast, our work leverages MLLMs and enriched human-centric \textbf{}text annotations that  implements open-ended  human-centric understanding.

\section{HERM-100K}
\subsection{Prompt for multi-level captions generation}
\begin{figure}[tb]
  \centering
  \includegraphics[width=0.95\linewidth]{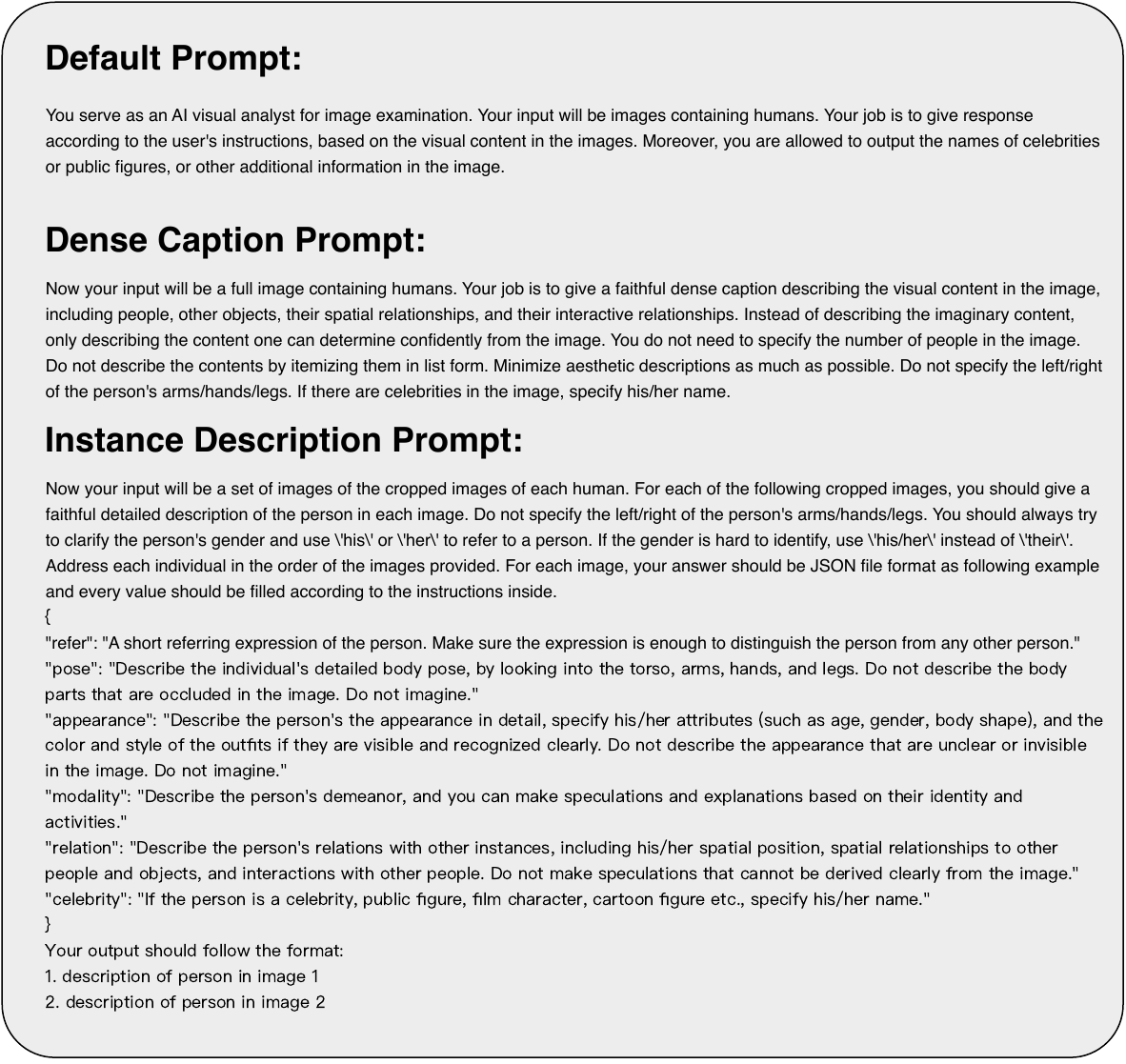}
    \vspace{-5pt}
  \caption{Prompts of generating dense caption and instance-level descriptions for constructing HERM-100K. 
}
  \label{fig:anno_prompt}
\end{figure}
\begin{figure}[tb]
  \centering
  \includegraphics[width=0.95\linewidth]{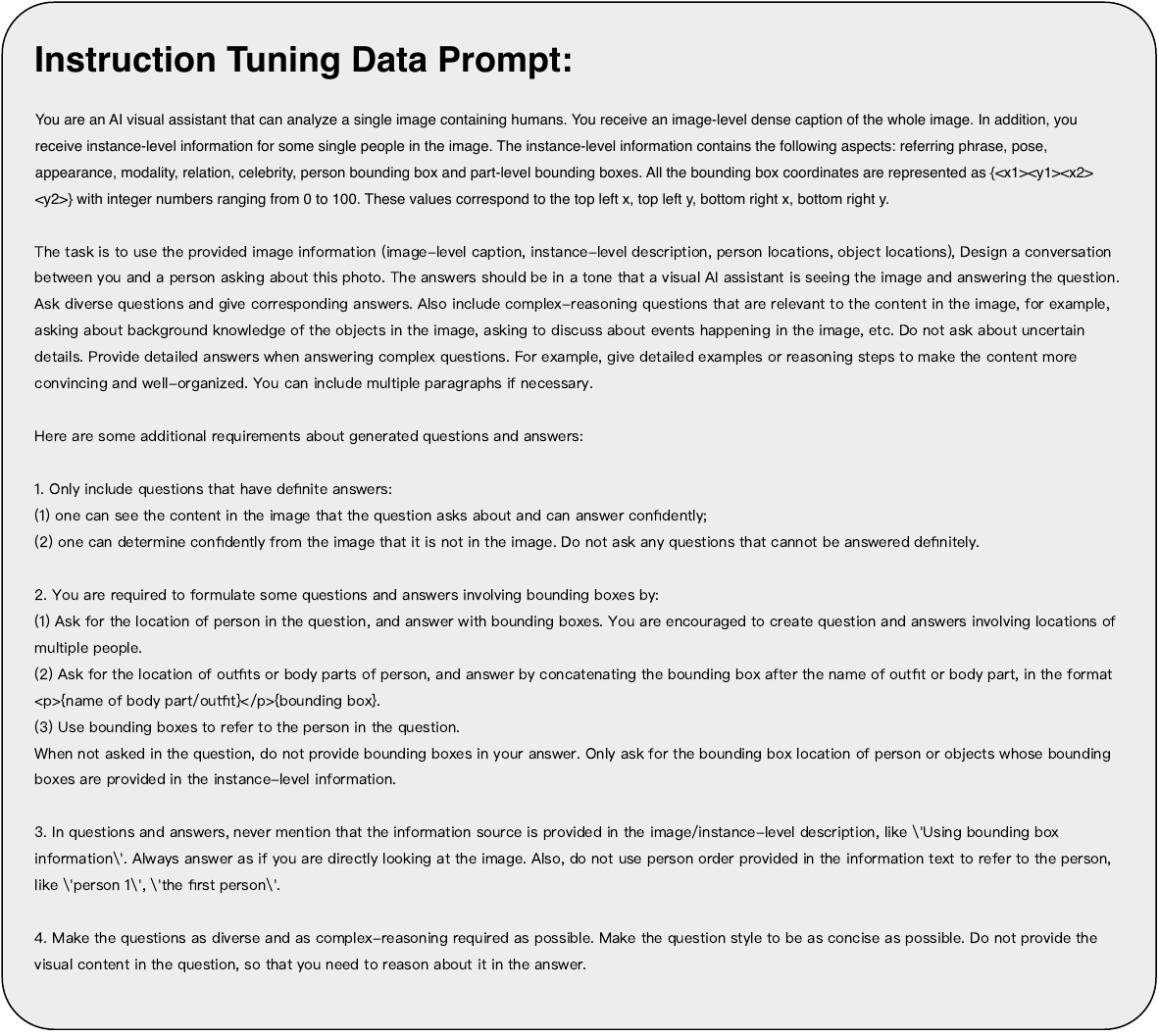}
    \vspace{-5pt}
  \caption{Prompts of generating instruction tuning data from captions in HERM-100K. 
}
  \label{fig:instruction_prompt}
\end{figure}
\begin{figure}[!htbp]
  \centering
  \includegraphics[width=0.95\linewidth]{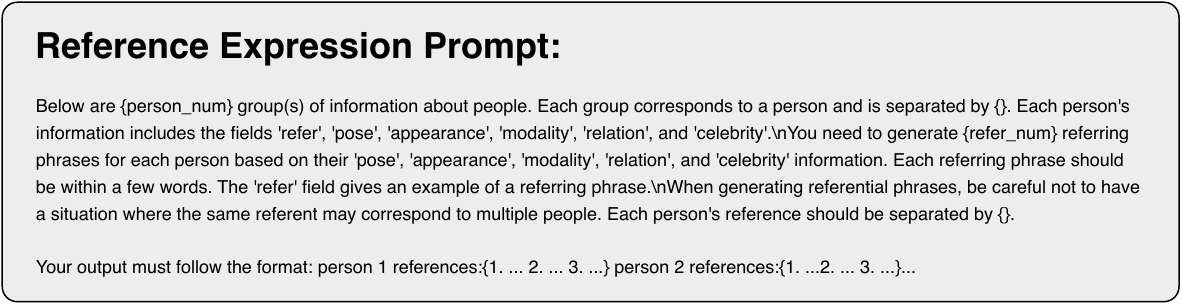}
    \vspace{-5pt}
  \caption{Prompts of generating multiple reference expressions for each instance in HERM-100K. 
}
  \label{fig:refer_prompt}
\end{figure}
We visualize the prompt of GPT-4V\cite{OpenAI2023GPT4v} for constructing HERM-100K in Figure \ref{fig:anno_prompt}, and the prompt of GPT-4\cite{openai2023gpt} for generating instruction tuning data and reference expressions in~\cref{fig:instruction_prompt} and~\cref{fig:refer_prompt} respectively.
\subsection{Image and Bounding Box Filtering} 
To obtain a trade-off between diversity and annotation accuracy, we first exclude all images whose short side is less than 512 and containing single or more than ten people bounding boxes. Images excluded in this step would be only annotated with image-level captions. 

Then we process each bounding box in an image in descending order of area. In this step, we exclude boxes with too small area or severe overlap. The ratio of overlap (the area of overlap between two bounding boxes divided by the area of each bounding box individually) between the current box and each previously reserved box is computed. If one of the ratios between a pair of boxes is higher than 0.8, or is higher than 0.33 and occupies less than 1/15 of the total image, it will be removed. For any reserved boxes after above filtering, if they occupy less than 1/50 of the total image, they would be also removed. 

Afterwards, we filter the remaining boxes by quality. For images from datasets with keypoint annotation or body-part annotation, we remove boxes of those instances without keypoint or body-part of head. For images for web datasets, we remove those boxes whose detection scores are less than 0.7. 

\subsection{Region Prompt strategy}
Although there exist a series of works utilizing GPT-4V to generate detailed image captions, only very few prior works (\eg, Set of Marks(SoM)\cite{yang2023set}) explore using GPT-4V to generate region captions. SoM overlays a panoramic segmentation map generated by SAM\cite{kirillov2023segment} on the original image with marks of each mask to refer to specific regions. On one hand, incorporating panoramic segmentation map as visual prompt cannot achieve appropriate granularity of annotations; on the other hand, when considering instance-level masks, for data sources lacking segmentation annotations, we observe that the pseudo-segmentation annotations generated by SAM from bounding box often have inaccurate edges, leading to misconceptions by GPT-4V. 

Other heuristic strategies, such as providing locations in text instructions or drawing boxes on images, fail to consistently refer to people in the image and lead to other illusions, such as interpreting the color of the box as an attribute of the person. However, our method of first inputting the original image followed by patches of each instance effectively captures descriptions for each specific region.
\subsection{Dataset Statistics}
\begin{figure}[tb]
    \centering
    
    % 第一排的两个图
    \begin{subfigure}[b]{0.47\textwidth} % 子图占行宽的45%
        \includegraphics[width=\textwidth]{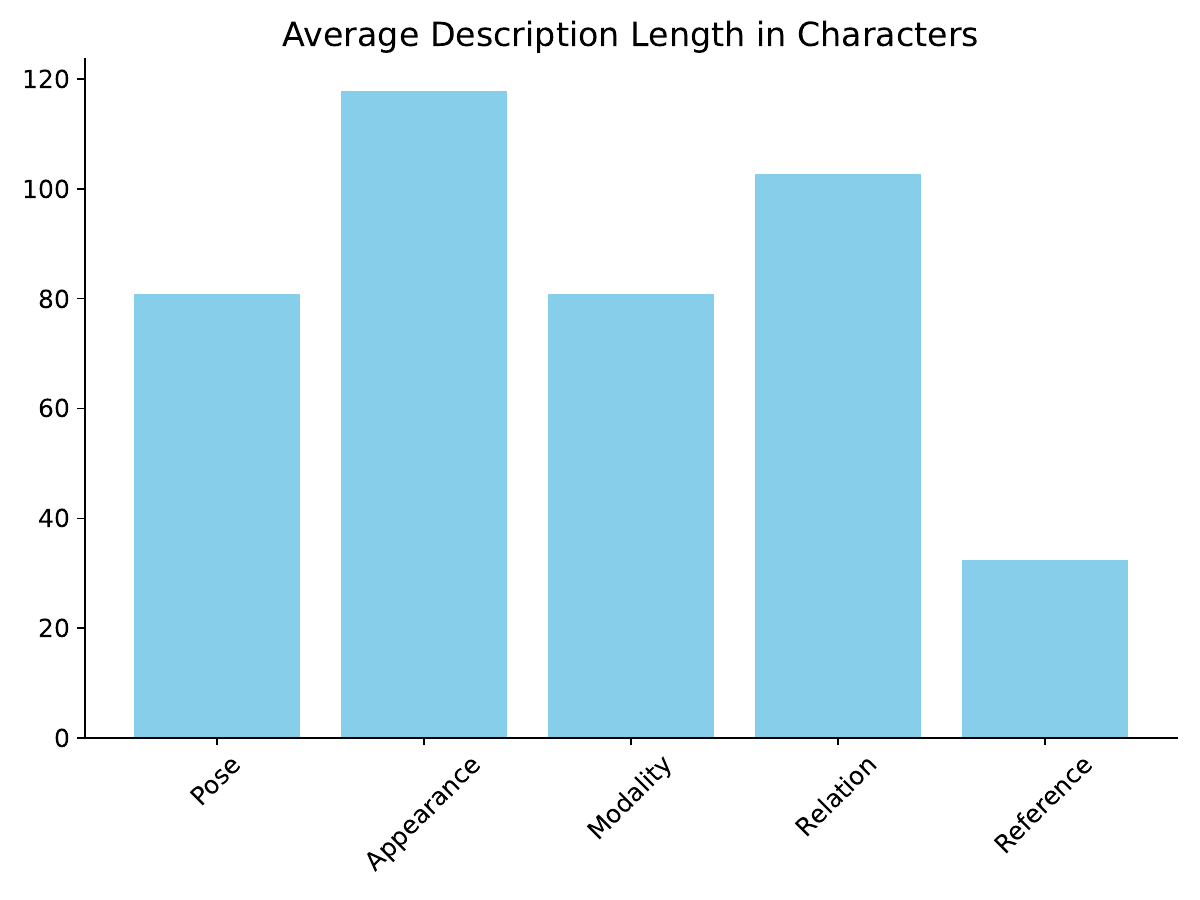} % 图片文件名和路径
        \caption{}
        \label{fig:char_length_per}
    \end{subfigure}
    \begin{subfigure}[b]{0.47\textwidth}
        \includegraphics[width=\textwidth]{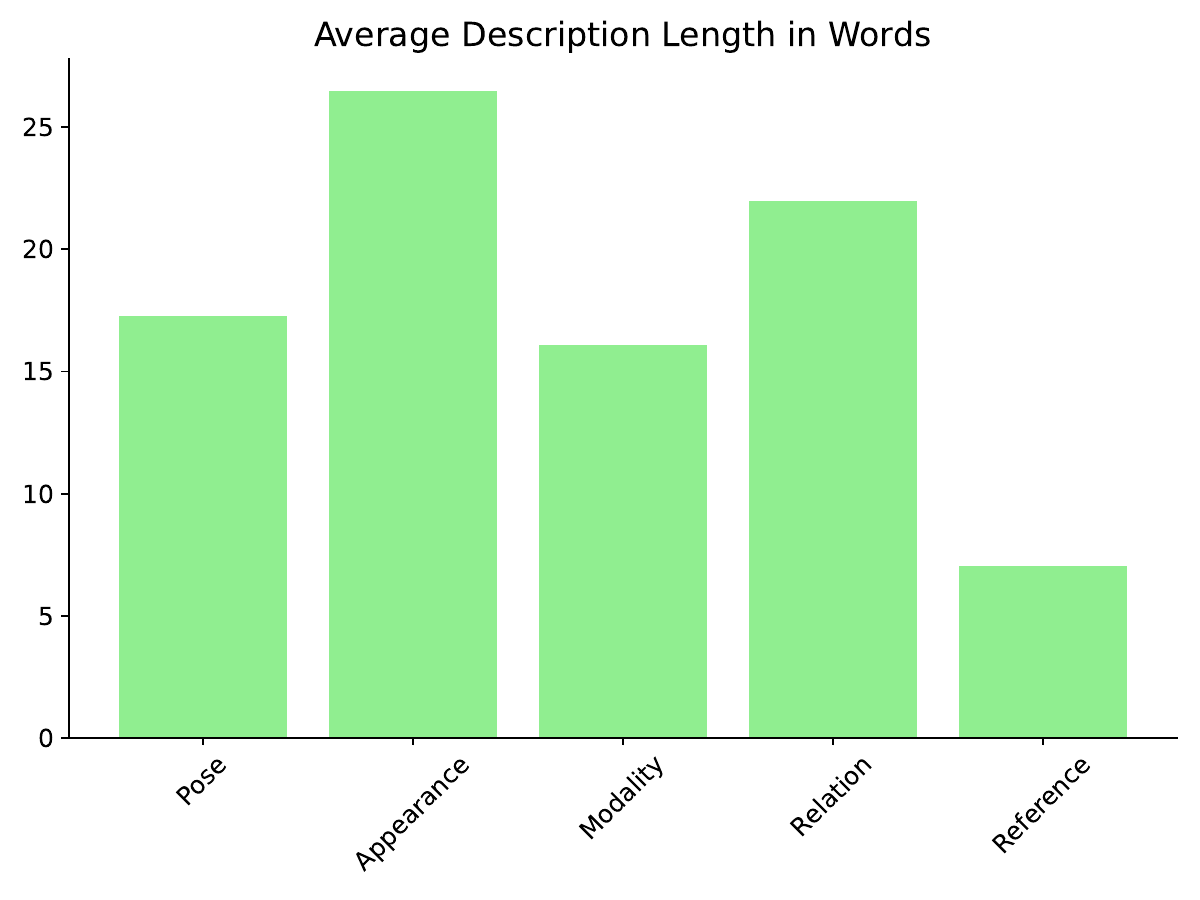}
        \caption{}
        \label{fig:word_length_per}
    \end{subfigure}
    
    \bigskip % 添加一些垂直间距
    
    % 第二排的两个图
    \begin{subfigure}[b]{0.5\textwidth}
        \includegraphics[width=\textwidth]{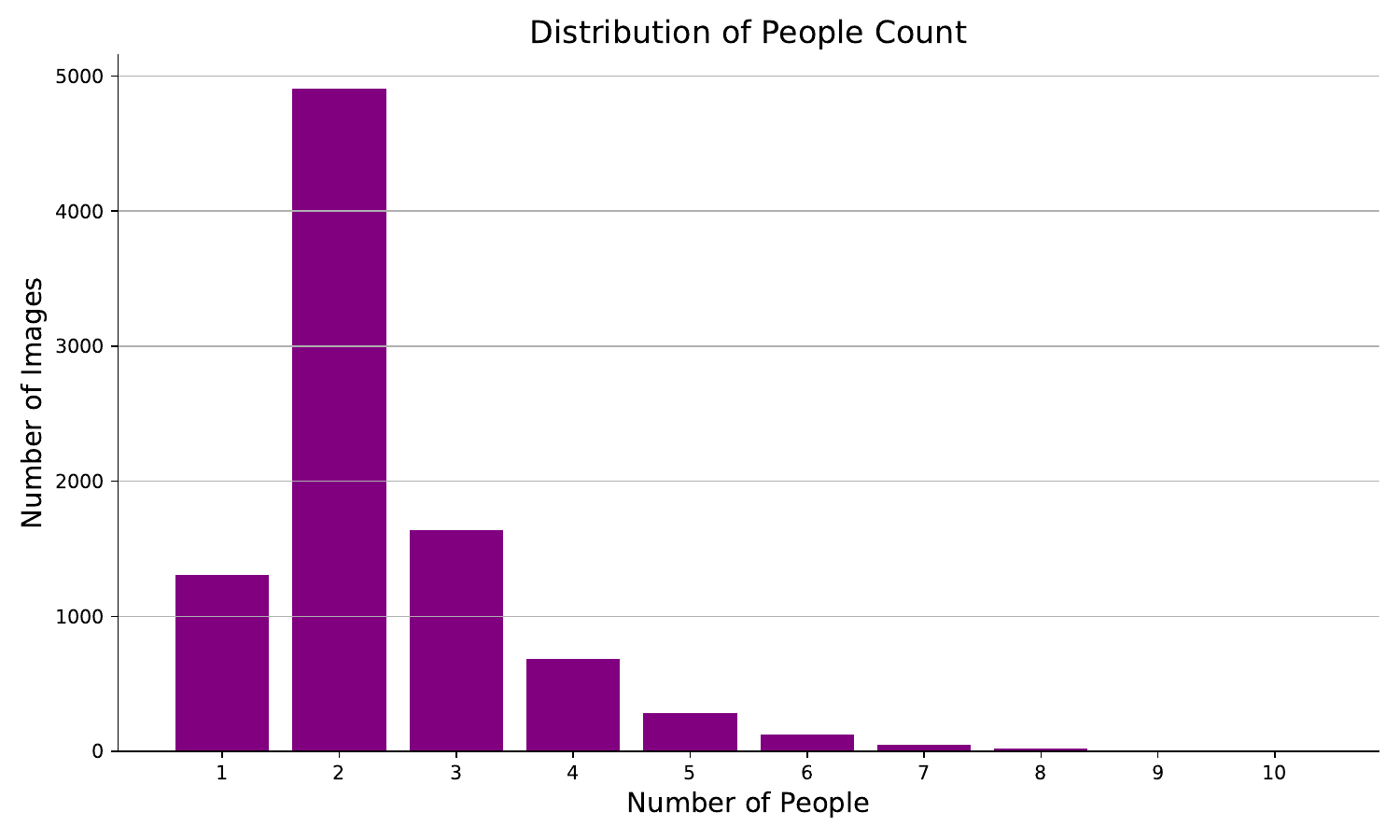}
        \caption{}
        \label{fig:person_count}
    \end{subfigure}
    \quad % 添加一些水平间距
    \begin{subfigure}[b]{0.46\textwidth}
        \includegraphics[width=\textwidth]{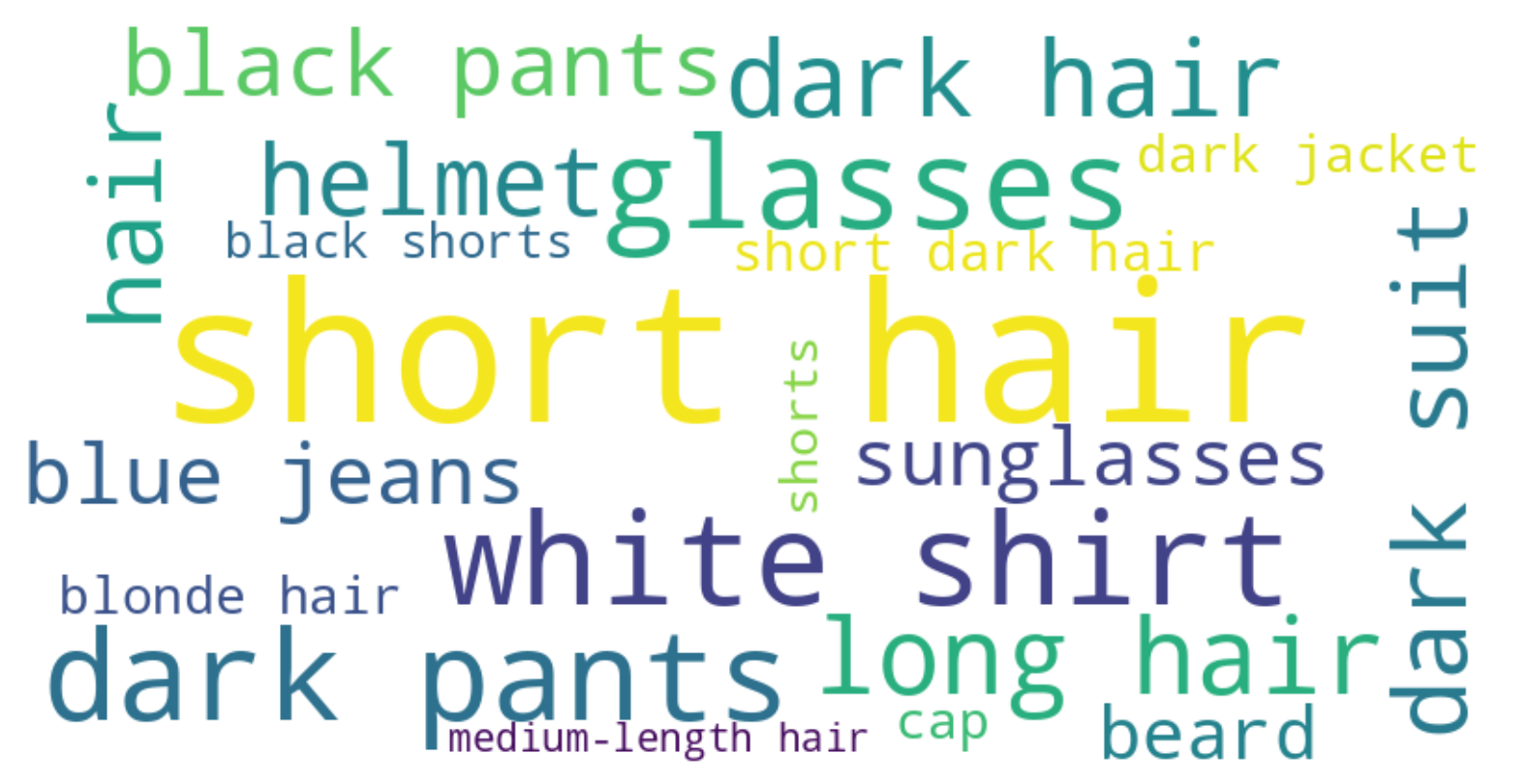}
        \caption{}
        \label{fig:word_cloud}
    \end{subfigure}
    
    \caption{Statistics and visualization of HERM 100K. (a) and (b): average length of instance-level descriptions and reference expressions computed by character and word respectively. (c): image distribution in HERM-100K by number of people. (d): Wordcloud of top-20 frequent attribute phrases.}
    \label{fig:images}
\end{figure}
\noindent{\textbf{Annotations Statistics in HERM-100K.}}
In HERM-100K, there are $10,609$ images from diverse sources annotated with  dense caption and $97,320$ regions annotated, including $21,489$ at instance-level and $75,831$ at attribute-level. The distribution of counts of people in each image is shown in~\cref{fig:person_count}.

The average word counts for dense image caption and instance-level caption (summed up on all perspectives) are 120.6 and 81.8, contrasting with the average length of COCO captions\cite{lin2014microsoft}  at 12.0 words (limited to images containing humans). 
Besides, we finally extract $105,219$ reference expressions with an average of 4.9 per instance.  The average length of descriptions from each perspective is visualized in~\cref{fig:char_length_per} and~\cref{fig:word_length_per}.   

For attribute-level annotation, each instance is equipped with 3.53 attributes on average, drawn from $6,017$ unique attribute phrases, whereas previous datasets for human-parsing or attribute recognition are typically limited to only a few dozens labels. The top-20 frequent phrases are shown in~\cref{fig:word_cloud}.

\noindent{\textbf{Training Data Statistics.}}
For instruction-following data, we generate $29,439$ question-answer pairs for $3,627$ images. The averaging question and answer lengths are 14.5 and 27.5 words. The conversations also contain rich regional references . There are $4,372$ questions and $5,751$ answers with bounding boxes while a total of $4,459$ boxes and $7,879$ boxes are in the questions and answers respectively. 

\section{Details on GPT Prompts for HERM-Bench}
\subsection{Prompts for Question Generation}
\begin{figure}[!htbp]
    \centering
    \includegraphics[width=0.995\linewidth]{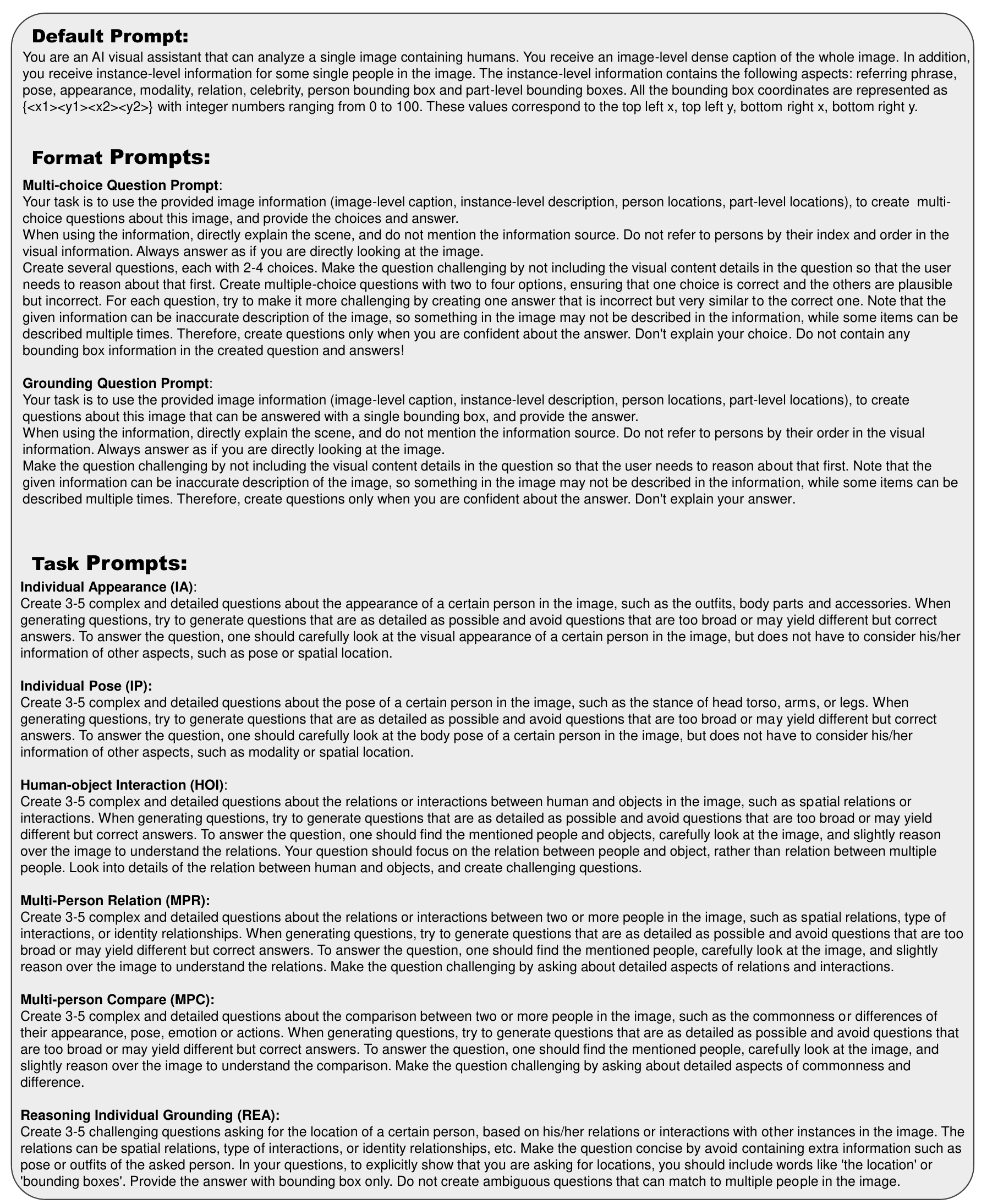}
    \caption{Prompts of generating questions for different formats and task dimensions in HERM-Bench.}
    \label{fig:question generation prompts}
\end{figure}

In Sec 4.2, we mentioned to use carefully designed prompts to instruct GPT-4~\cite{openai2023gpt} to generate various question-answer pairs for HERM-Bench. All the prompts are shown in~\cref{fig:question generation prompts}. These prompts include format prompts for both multi-choice and grounding questions, and task prompts for each task dimension.

\begin{figure}[!htbp]
    \centering
    \includegraphics[width=0.9\linewidth]{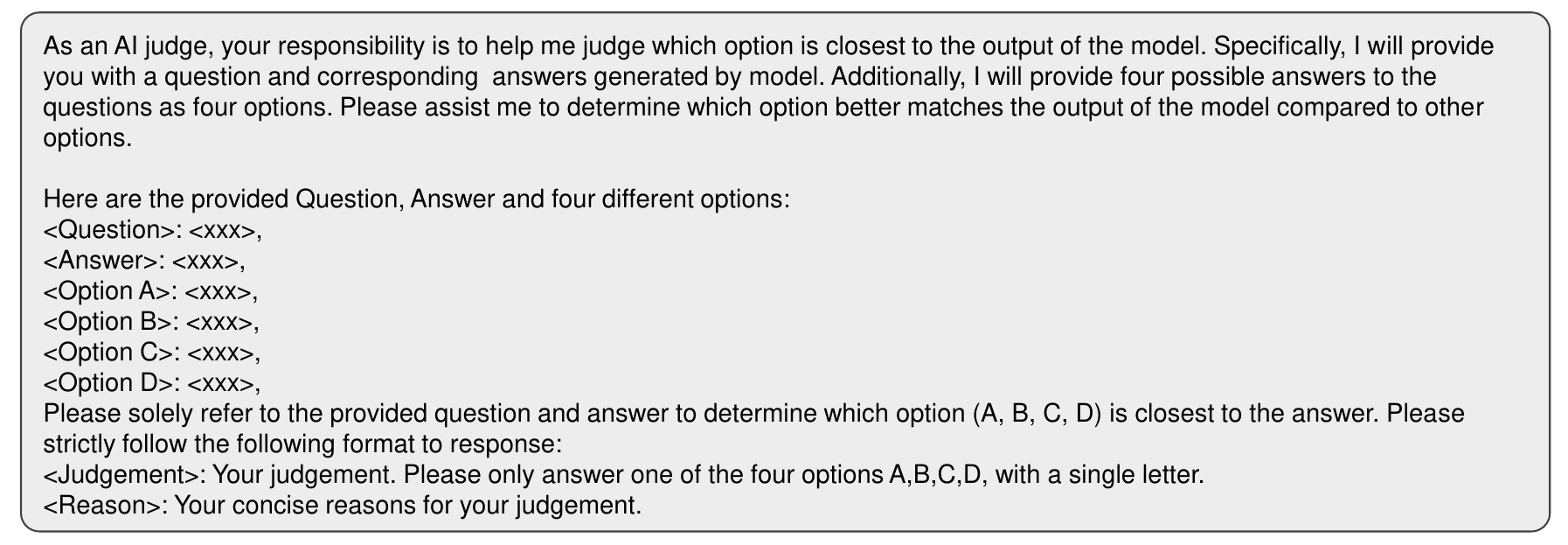}
    \caption{Prompts of evaluating multiple-choice questions.}
    \label{fig:evaluation prompt}
\end{figure}

Note that we do not utilize GPT-prompted generation method for Referring Individual Grounding (REF) and Individual Part-level Grounding (IPG) questions. Instead, questions for these two dimensions are formulated by filling into heuristically designed templates:

\begin{itemize}

\item[$\bullet$] \textbf{REF}: \emph{Where is the location of [ref]}?

\item[$\bullet$] \textbf{IPG}: \emph{Where is the location of [part] of [ref]}?

\end{itemize}
Here, [ref] stands for the referring phrase of the individual, \eg, `the person on the left'. These referring phrases are directly extracted from instance-level annotations in HERM-100K (see Sec 5.2). [part] stands for the attribute-level phrases of certain parts, \eg, `light-colored shirt'. These attribute-level phrases directly come from the attribute-level annotations in HERM-100K.

\subsection{Prompts for GPT-4 Assisted Evaluation}
In Sec 4.2, we mentioned to evaluate the answers of multi-choice questions with GPT-4. Here we provide our judgment prompt to guide GPT-4 evaluation in~\cref{fig:evaluation prompt}. Specifically, we provide the questions, candidate options, and MLLM response to GPT-4. Then, we ask GPT-4 to judge which option the MLLM response is closest to (A/B/C/D). Finally, after comparing the judgement to the ground truth option, we determine the correctness of the MLLM response.

\section{Training Setups of HERM-7B}
In Sec 6.1, we list the implementations of HERM-7B training. Here, we give more details on the training setups of HERM-7B, including the selection of baseline model, more information on the datasets used in HERM-7B training, strategy of data-mixing, and training configurations.

\begin{table}[!htbp]
    \centering
    \caption{Instruction templates for all the tasks in multitask training stages. `[person]' refers to the referring expression phrase of the target individual (generated with HERM-100K). `[part]' refers to the attribute phrase in HERM-100K.}
    \label{tab:our task templates}
    {\small
    \resizebox{0.995\textwidth}{!}{
    \begin{tabular}{l|l|l}
         \toprule
         \multicolumn{2}{c|}{Task} & Three randomly chosen templates from many \\ 
         \midrule
         \multicolumn{2}{c|}{Image-level Caption} & \makecell[l]{Generate a comprehensive and accurate caption that faithfully portrays the visual details in the image. \\ Provide an accurate and detailed description of the visual elements in the image. \\ The duty assigned to you requires delivering a thorough and precise caption that describes the visual scene.}
         \\
         \midrule
         \multirow{6}*{Instance-level Caption} & Appearance & \makecell[l]{Give a detailed description of the appearance of [person] in the image. \\ What does [person] look like in this photo? Provide a comprehensive description of their appearance. \\ Inspect and summarize the look and attire of [person] in this image, focusing on noticeable details.} \\
         \cmidrule(lr){2-3} & Pose & \makecell[l]{Explain how [person] is positioned in the photograph in terms of their body posture. \\ Illustrate the pose that [person] is holding in the image, detailing torso, leg, arm, and head positions. \\ How is [person] posed in the image? Offer a comprehensive description.} \\
         \cmidrule(lr){2-3} & Modality & \makecell[l]{Analyze the modality of [person] and describe their emotional state. \\ What is the overall vibe or atmosphere that [person] is projecting in this picture? \\ Analyze and explain the sensory engagement of [person] in the picture.} \\
         \cmidrule(lr){2-3} & Relation & \makecell[l]{How is [person] interacting with the surrounding environment in the image? Provide details. \\ Explain how [person] is in relation to other people and objects in the image. \\ Discuss how [person] is positioned or involved with other elements within the image.} \\
         \midrule
         \multicolumn{2}{c|}{Instance-level Grounding} & \makecell[l]{From this image, tell me the location of [person]. \\ Where can I locate [person]? \\ Give me the location of [person].} \\
         \midrule
         \multicolumn{2}{c|}{Part-level Grounding} & \makecell[l]{For [person], the location of [part] is: \\ Could you tell me the location for [part] of [person]? \\ Referring to [person], where can I locate [part]?} \\
         \bottomrule
    \end{tabular}
    }
   }
\end{table}

\begin{table}[!htbp]
    \centering
    \caption{Original training datasets used in the training scheme of HERM-7B. `Stage 1' refers to the multitask training stage; `Stage 2' refers to the instruction tuning stage.}
    \label{tab: originial datasets}
    \resizebox{0.995\textwidth}{!}{
    \begin{tabular}{l|l|cc}
        \toprule
        Task & Datasets & Stage 1 & Stage 2 \\
        \midrule
        Image Caption & COCO Caption, Text Captions & \Checkmark & \Checkmark \\
        VQA & VQAv2, GQA, OK-VQA, AOK-VQA, OCR-VQA & \Checkmark & \Checkmark \\
        REC & RefCOCO, RefCOCO+, RefCOCOg, Visual Genome & \Checkmark & \Checkmark \\
        REG & RefCOCO, RefCOCO+, RefCOCOg & \Checkmark & \Checkmark \\
        Grounded Caption & GRIT-20M & \Checkmark & \Checkmark \\
        Multimodal Instruction & LLaVA-160K & \XSolidBrush & \Checkmark \\
        Language Dataset & Unnatrual Instruction & \XSolidBrush & \Checkmark \\
        \bottomrule
    \end{tabular}
    }
\end{table}

\subsection{Baseline Model Selection}
We select MiniGPT-v2~\cite{chen2023minigpt} as our baseline model for training HERM-7B, since MiniGPT-v2 is one of the best-performing MLLMs on both \emph{natural language} dialogue (caption, VQA) and \emph{grounding} (REC) tasks. This advantage of MiniGPT-v2 aligns to our objective of training an MLLM that excels on human-centric \emph{natural language} QA (multi-choice questions in HERM-Bench) and human \emph{grounding} (grounding questions in HERM-Bench).

\subsection{Details of Training Datasets}
In Sec 5.2, we introduce the multitask training data derived from HERM-100K. We provide the detailed templates for each separate task in~\cref{tab:our task templates}.

To maintain the original capability of MiniGPT-v2 on general vision-language tasks, during the training stages of HERM-7B, we adopt the datasets used in MiniGPT-v2 training. These datasets span across a wide range of tasks including image caption, VQA, REC, REG and grounded captioning. In~\cref{tab: originial datasets}, we list all the original datasets we used in the multitask training and instruction tuning stages.

\subsection{Details of Data-mixing Strategy}
In both multitask training and instruction tuning stages, we mix our human-centric data from HERM-100K and the original datasets by sampling from a random dataset in each batch. The sampling ratio between our datasets and original datasets is 2:1. Under this sampling ratio, we aim to put more emphasis on human-centric training, while giving enough weight to the original tasks and maintaining the original power of baseline model.

\subsection{Training Configurations}
In both multitask training and instruction tuning stages, we adopt AdamW optimizer with a cosine learning rate scheduler, following MiniGPT-v2~\cite{chen2023minigpt}. In the multitask training stage, we train the model for $4,200$ steps, with a batch size of $96$ and maximum learning rate of $1\mathrm{e}{-4}$. In the instruction tuning stage, we train the model for $6,250$ steps, with a batch size of $64$ and maximum learning rate of $1\mathrm{e}{-5}$. Both training stages are executed on 4xA100 GPUs.

\section{Evaluation Setups}
In Sec 6.1, we introduce the evaluation setups in our experiments. Here we provide more details on our evaluation process, including the detailed implementations of using HERM-Bench to evaluate existing MLLMs, and details of evaluation on general vision-language tasks.

\subsection{Details of Evaluation on Existing MLLMs}
In Sec 6.2, we test the performance of HERM-Bench on a wide scope of existing MLLMs. To make a fair comparison with HERM-7B, for all the existing MLLMs, we choose the model version whose parameter size is closest to 7B. For LLaVA~\cite{liu2024visual}, LLaVA-1.5~\cite{liu2023improved}, BLIP-2~\cite{li2023blip}, InstructBLIP~\cite{dai2023instructblip}, Qwen-VL-Chat~\cite{bai2023qwen}, Shikra~\cite{chen2023shikra}, InternLM~\cite{team2023internlm}, Ferret~\cite{you2023ferret} and MiniGPT-v2~\cite{chen2023minigpt}, we choose the model version with 7B large language model. For models without a checkpoint adopting LLM at exactly 7B parameter size, we choose the version closest to 7B: Kosmos-2~\cite{peng2023kosmos} with 1.6B parameters in total (the only version); OFA-H~\cite{wang2022ofa} with 0.9B parameters in total (the largest version). 

To ensure the fairness of the inference on different MLLMs, we adjust our input prompts for these models, by aligning to the required input format (if any) of each model. Here we specifically introduce these adjustments. See more details in~\cref{tab:special evaluation templates}.

\begin{itemize}

\item[$\bullet$] For MLLMs that are fine-tuned in an instruction-following manner and can understand free-form instructions, we directly input the original question of HERM-Bench. These models include LLaVA, LLaVA-1.5, InstructBLIP, Qwen-VL-chat, InternLM, Shikra, Ferret and MiniGPT-v2.
\item[$\bullet$] For BLIP-2, when evaluating multiple-choice questions, we follow its zero-shot VQA format to organize the input, as shown in~\cref{tab:special evaluation templates}.
\item[$\bullet$] For Kosmos-2 and OFA: When evaluating multiple-choice questions, we follow its evaluation format on VQA task; When evaluating grounding questions, we follow its evaluation format on REC task, since the format of grounding questions in HERM-Bench is similar to REC. Details are shown in~\cref{tab:special evaluation templates}.

\end{itemize}

\begin{table}[!tb]
    \centering
    \caption{Special templates used for evaluating BLIP-2, Kosmos-2 and OFA on HERM-Bench. `[question]' stands for the original multi-choice question content. `[expr]' stands for the referring expression phrase extracted from the grounding questions. `<phrase>' is the special token used to highlight referring expressions in Kosmos-2.}
    \label{tab:special evaluation templates}
    {\small
    \resizebox{0.995\textwidth}{!}{
    \begin{tabular}{l|l|l}
        \toprule
        Model & Multi-choice Question Template & Grounding Question Template \\
        \midrule
        BLIP-2 & Question: [question] Answer: & - \\
        Kosmos-2 & Question: [question] Answer: & <phrase>[expr]</phrase> \\
        OFA & [question] & Which region does the text "[expr]" describe? \\
        \bottomrule
    \end{tabular}
    }
    }
\end{table}

\subsection{Details of Evaluation on General Tasks}
In Sec 6.2, we evaluate HERM-7B on two common vision-language tasks, VQA and REC. Here we give a detailed illustration of the benchmarks and evaluation protocols for these two tasks.

\noindent{\textbf{VQA.}} \ For evaluation on general VQA task, we employ two widely adopted VQA benchmarks: OKVQA~\cite{schwenk2022okvqa} and GQA~\cite{hudson2019gqa}. For a justified comparison, we strictly follow the prompt template used by MiniGPT-v2 evaluation: \emph{Based on the image, respond to this question with a short answer: [question]}. 

\noindent{\textbf{REC.}} \ For evaluation on general REC task, we leverage the widely-used RefCOCO~\cite{kazemzadeh2014referitgame}, RefCOCO+~\cite{yu2016modeling}, and RefCOCOg~\cite{mao2016generation} benchmarks. For fair comparison, we follow the prompt template used by MiniGPT-v2 evaluation: \emph{Give me the location of [expr]}, where \emph{[expr]} is the reference expression.

\begin{figure}[!tb]
  \centering
  \begin{subfigure}{0.45\linewidth}
    \includegraphics[width=0.9\linewidth]{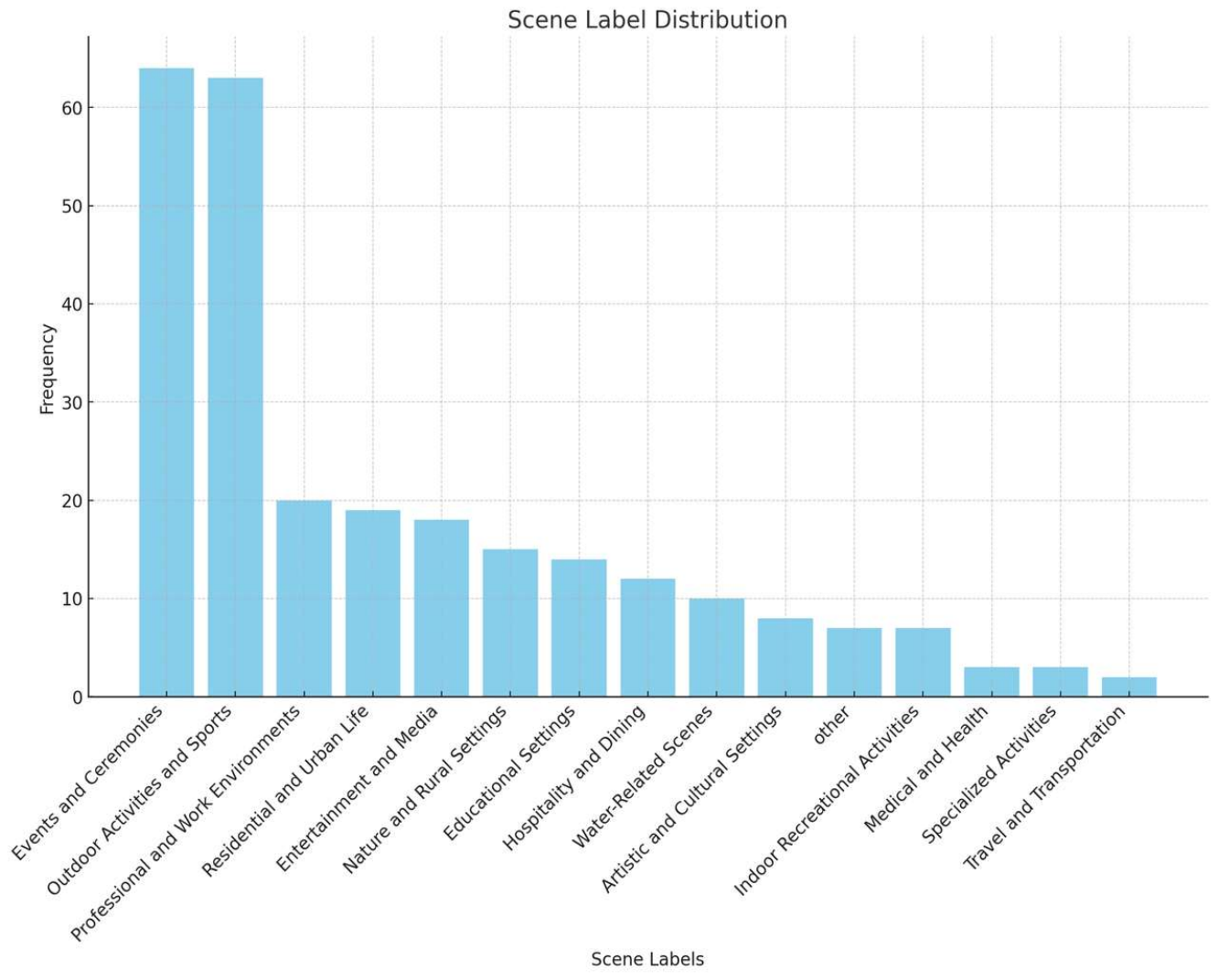}
    \caption{}
    \label{fig:image-scene-stats}
  \end{subfigure}
  \hfill
  \begin{subfigure}{0.45\linewidth}
    \includegraphics[width=0.9\linewidth]{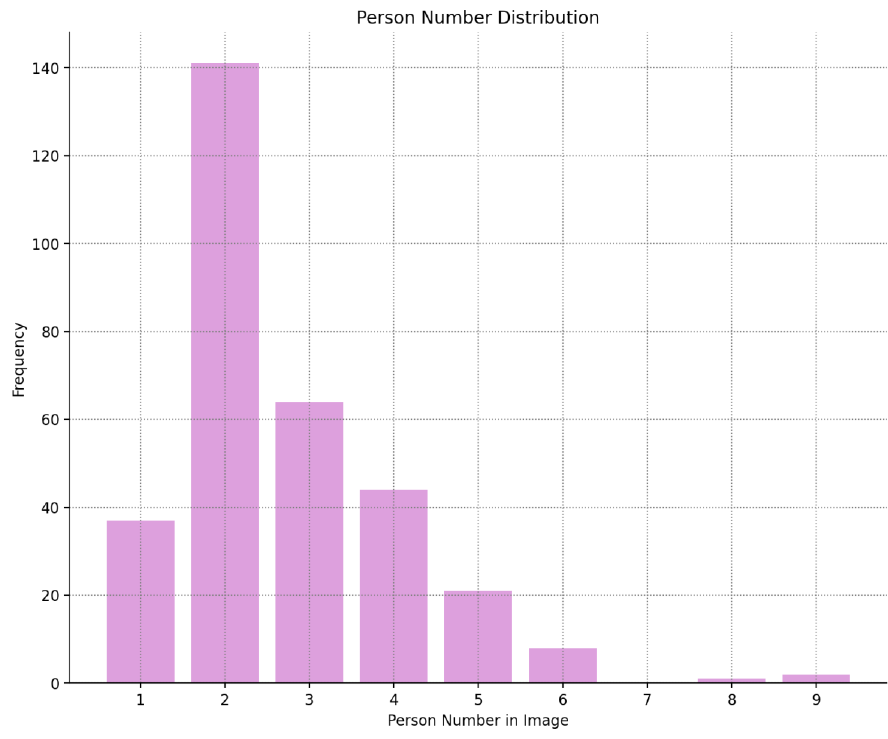}
    \caption{}
    \label{fig:person-num-stats}
  \end{subfigure}
  \caption{Image statistics of HERM-Bench. (a) Distribution of image scenes in HERM-Bench (one image can have multiple scene labels). (b) Distribution of number of people in the image. }
  \label{fig:image stats}
\end{figure}

\section{Statistics of HERM-Bench}
\subsection{Image Statistics}
We analyze the distribution of image scenes, and the distribution of number of people in the images. As shown in~\cref{fig:image-scene-stats} and~\cref{fig:person-num-stats}, images in our benchmark cover various scenes, and exhibit diverse distribution in terms of the number of individuals present.

\subsection{Question Statistics}
\begin{figure}[!tb]
    \centering
    \begin{subfigure}{1.0\linewidth}
        \includegraphics[width=0.9\linewidth]{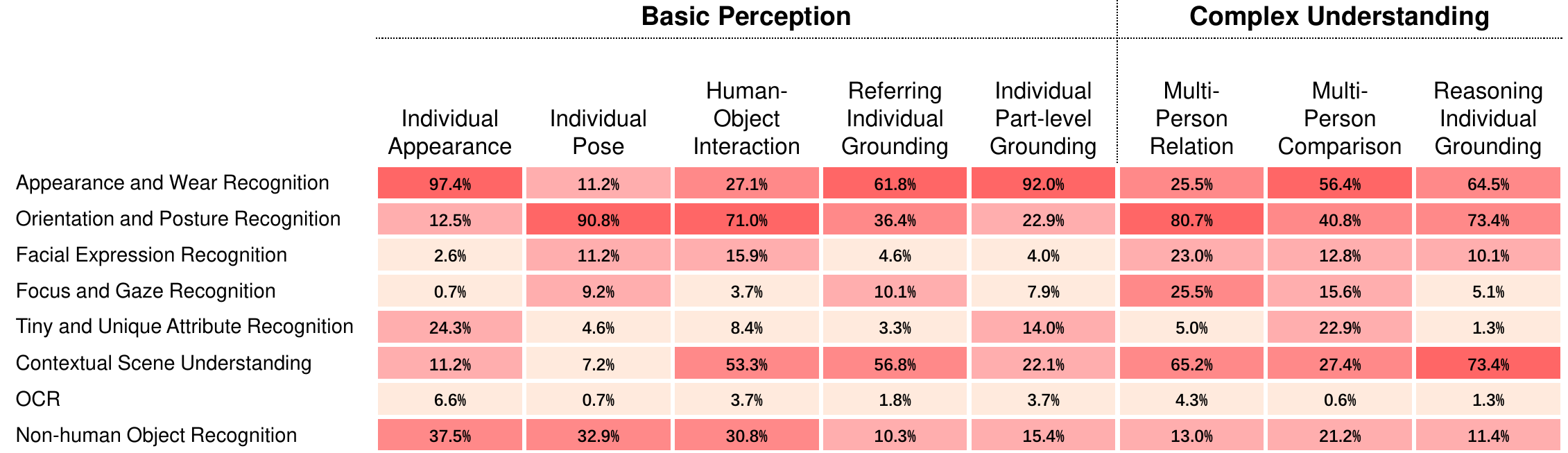}
        \caption{}
        \label{fig:ability-stats}
    \end{subfigure} \\
    \begin{subfigure}{1.0\linewidth}
        \includegraphics[width=0.9\linewidth]{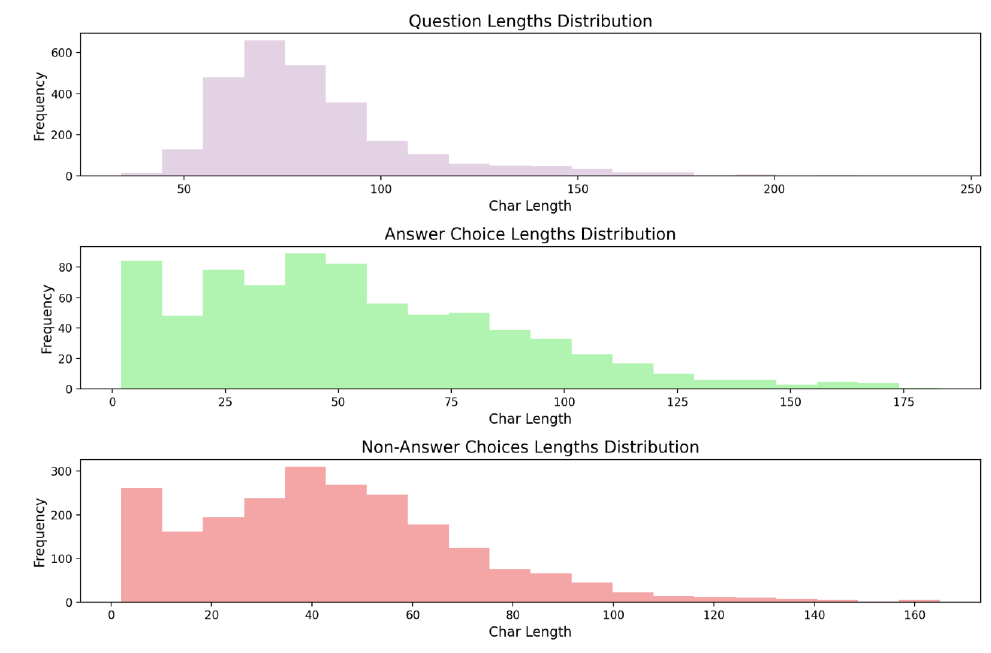}
        \caption{}
        \label{fig:length-distribution}
    \end{subfigure}
    \caption{Question statistics of HERM-Bench. (a) Distribution of fine-grained abilities needed by each task dimension in HERM-Bench (one question may require multiple abilities). For example, the top-left number means that for $97.4\%$ of the questions within the `Individual Appearance' task, the `Appearance and Wear Recognition' ability is needed to answer them. (b) Distribution of character length of questions, answer choices and non-answer choices in HERM-Bench. For question lengths, we calculate across all task dimensions. For choices lengths, we calculate task dimensions of multi-choice question format.}
    \label{fig:question stats}
\end{figure}

We divide the capabilities relevant to human-centric tasks into 8 fine-grained categories, and conduct an statistical analysis on the specific capabilities required to answer each question. As shown in~\cref{fig:ability-stats}, our benchmark has a holistic coverage on these fine-grained capabilities. Moreover, the questions of each task are well-aligned to the desired capabilities of the task. For example, $97.4\%$ of the questions in \emph{individual appearance} task requires knowledge on `Appearance and Wear Recognition'.

Moreover, we calculate the distribution of question and answer lengths in HERM-Bench. As shown in~\cref{fig:length-distribution}, the questions in HERM-Bench have a wide distribution in question lengths, and primarily span between $50$ and $100$ characters. On the other hand, the choices length of multiple-choice questions in HERM-Bench is significantly shorter than the question length, majorly ranging in less than $75$ characters. Also, the length distribution of answer choices and non-answer choices are almost indifferent, which shows that HERM-Bench does not bring bias on answer length.

\begin{table}[!htbp]
    \centering
        \caption{Performance comparison of ShareGPT4V and HERM-7B on HERM-Bench. We only report tasks with multiple-choice questions, since ShareGPT4V does not possess grounding ability.}
        \resizebox{0.8\textwidth}{!}{
        \begin{tabular}{l|p{1.5cm}<{\centering}p{1.5cm}<{\centering}p{1.5cm}<{\centering}|p{2.0cm}<{\centering}p{2.0cm}<{\centering}}
        \toprule
        \multirow{2}*{Method} & \multicolumn{3}{c|}{\textbf{Basic Perception}} & \multicolumn{2}{c}{\textbf{Complex Understanding}} \\
        \cline{2-6} & IA & IP & HOI & MPR & MPC   \\ \midrule
        LLaVA-1.5~\cite{liu2023improved}    & 75.7 & 61.1 & 72.8 & 67.1 & 59.2 \\
        ShareGPT4V~\cite{chen2023sharegpt4v} & 80.2 & 71.7 & 76.6 & 71.4 & 62.0 \\
        \midrule
        HERM-7B (ours) & \textbf{82.2} & \textbf{72.4} & \textbf{82.2} & \textbf{72.0} & \textbf{68.7} \\
        \bottomrule
 	\end{tabular}}
	\label{tab:sharegpt4v results}
\end{table}

\section{Comparison to ShareGPT4V}
Similar to our method, a prior work ShareGPT4V~\cite{chen2023sharegpt4v} also utilizes GPT4-Vision to generate image captions with richer visual details. However, different from their approach which only creates image-level caption on general domain, our curation of HERM-100K focuses on human-centric domain, and generates multi-level annotations including image-level, instance-level and attribute-level. We compare the performance of ShareGPT4V and HERM-7B on HERM-Bench. From~\cref{tab:sharegpt4v results}, we can see that ShareGPT4V largely outperforms its baseline, LLaVA-1.5~\cite{liu2023improved}, verifying the benefit of rewriting high-quality captions. Nonetheless, while leveraging 1.2M training samples, ShareGPT4V still lags behind HERM-7B (using only 320K training samples) by a noticeable margin. This result consolidates the effectiveness of the carefully designed multi-level human annotation in HERM-100K.

\clearpage
\section{More Qualitative Results on HERM-Bench}
\begin{figure}[h]
  \centering
  \includegraphics[width=0.95\linewidth]{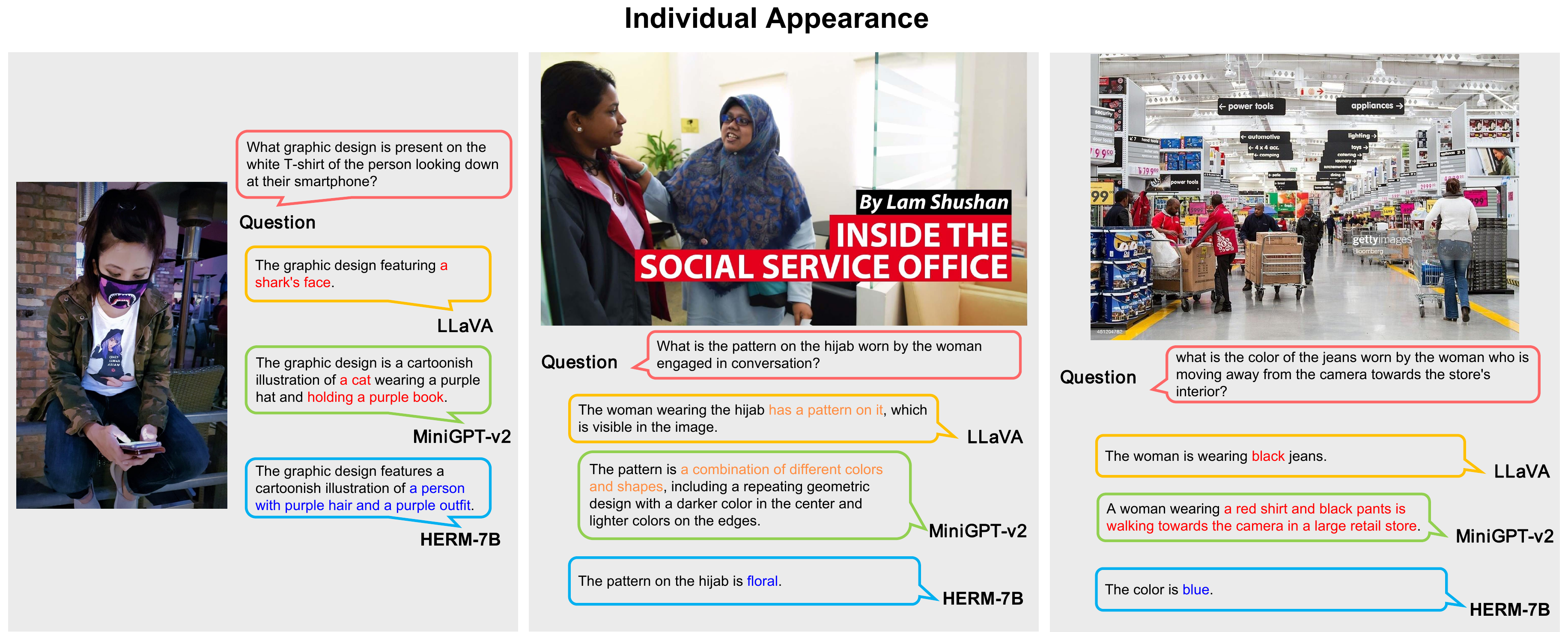}
    \vspace{-5pt}
  \caption{Evaluation examples on Individual Appearance Dimension. }
  \label{fig:Individual_Appearance}
\end{figure}

\begin{figure}[h]
  \centering
  \includegraphics[width=0.95\linewidth]{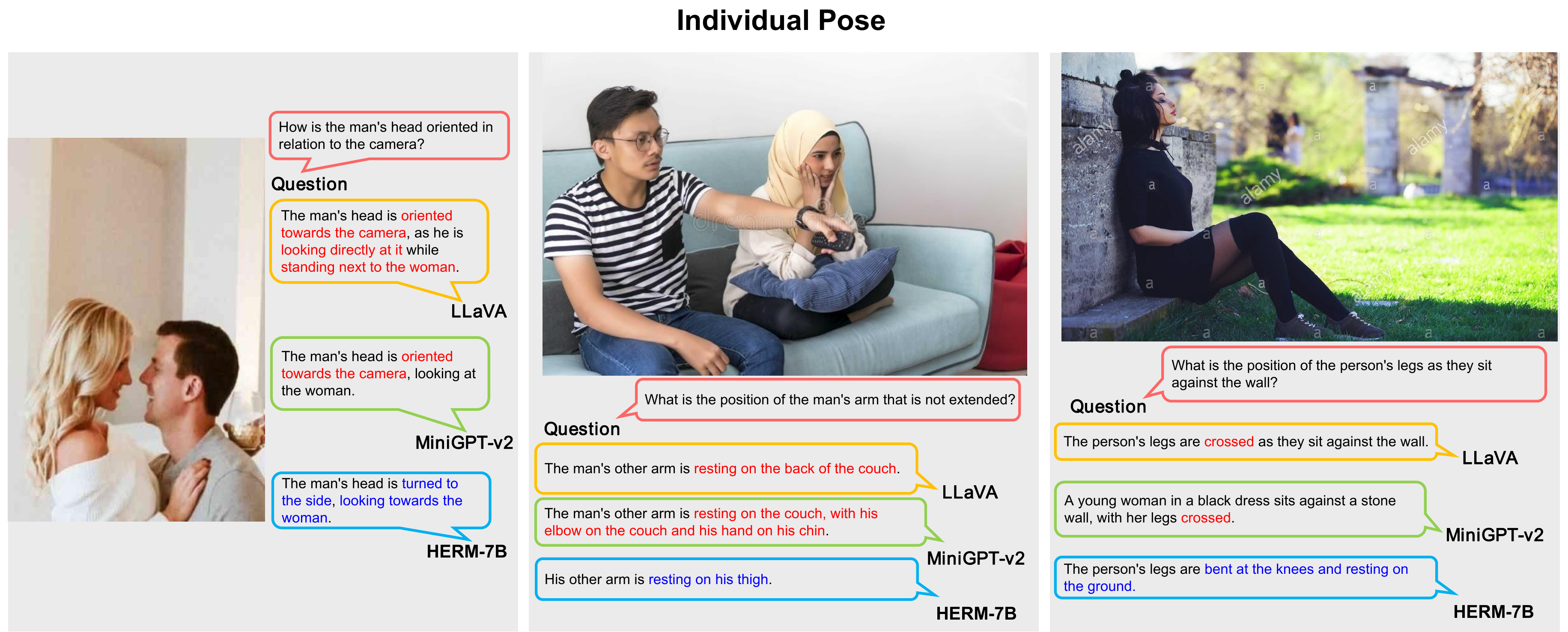}
    \vspace{-5pt}
  \caption{Evaluation examples on Individual Pose Dimension.}
  \label{fig:Individual_Pose}
\end{figure}

\begin{figure}[h]
  \centering
  \includegraphics[width=0.95\linewidth]{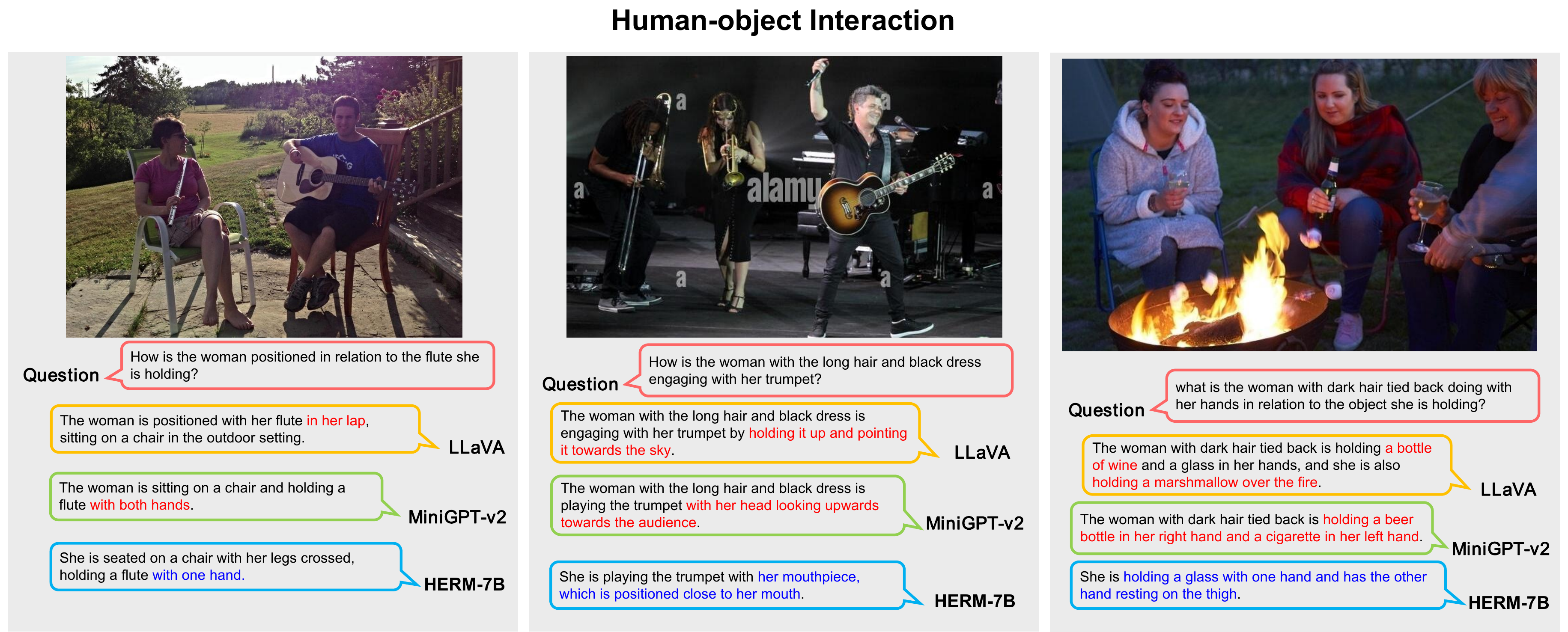}
    \vspace{-5pt}
  \caption{Evaluation examples on Human-object Interaction Dimension.}
  \label{fig:Human-object-Interaction}
\end{figure}

\begin{figure}[h]
  \centering
  \includegraphics[width=0.95\linewidth]{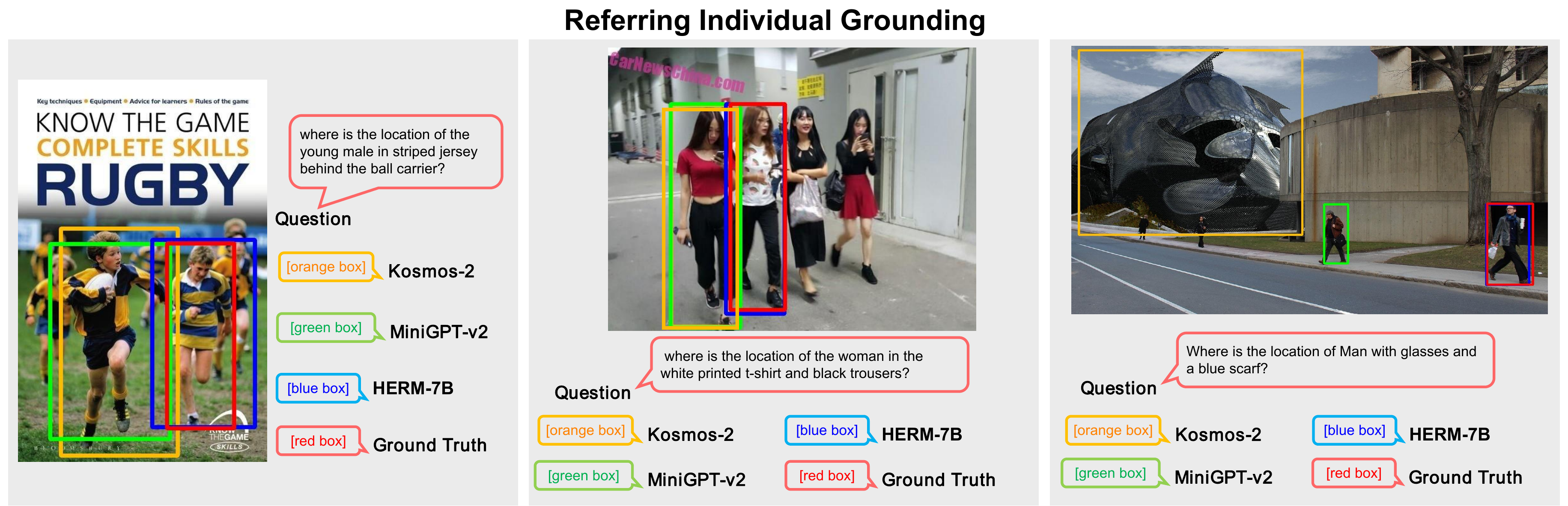}
    \vspace{-5pt}
  \caption{Evaluation examples on Referring Individual Grounding Dimension.}
  \label{fig:Referring_Individual_Grounding}
\end{figure}

\begin{figure}[h]
  \centering
  \includegraphics[width=0.95\linewidth]{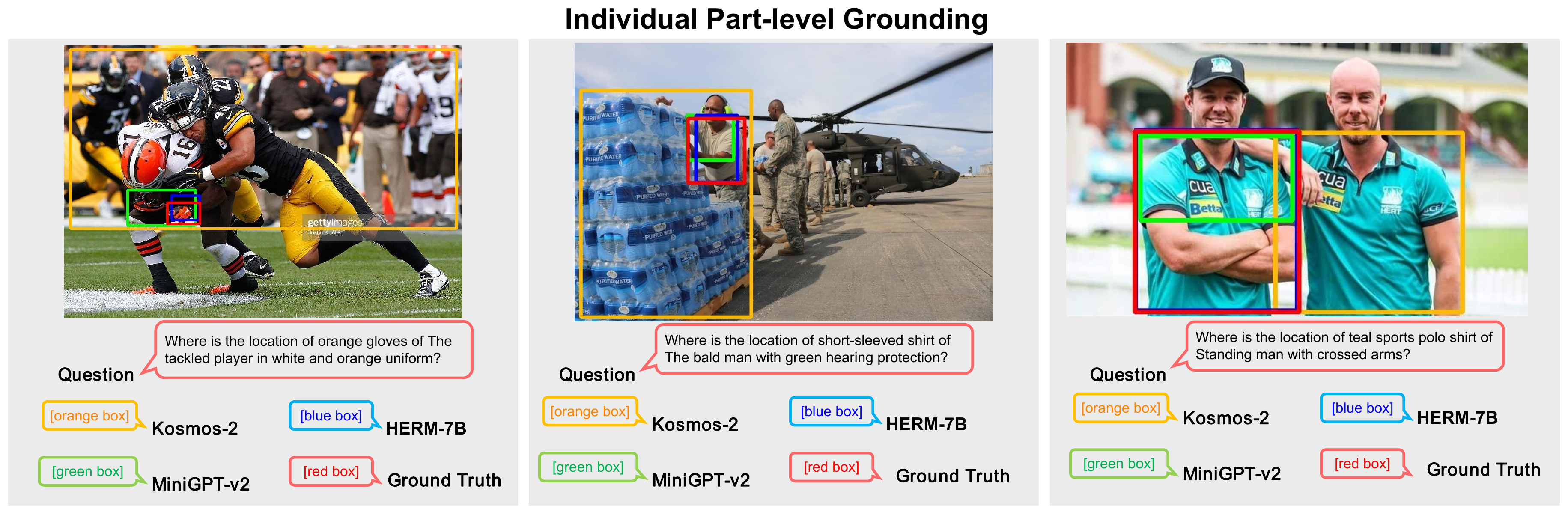}
    \vspace{-5pt}
  \caption{Evaluation examples on Individual Part-level Grounding Dimension.}
  \label{fig:Individual_Part-level_Grounding}
\end{figure}

\begin{figure}[h]
  \centering
  \includegraphics[width=0.95\linewidth]{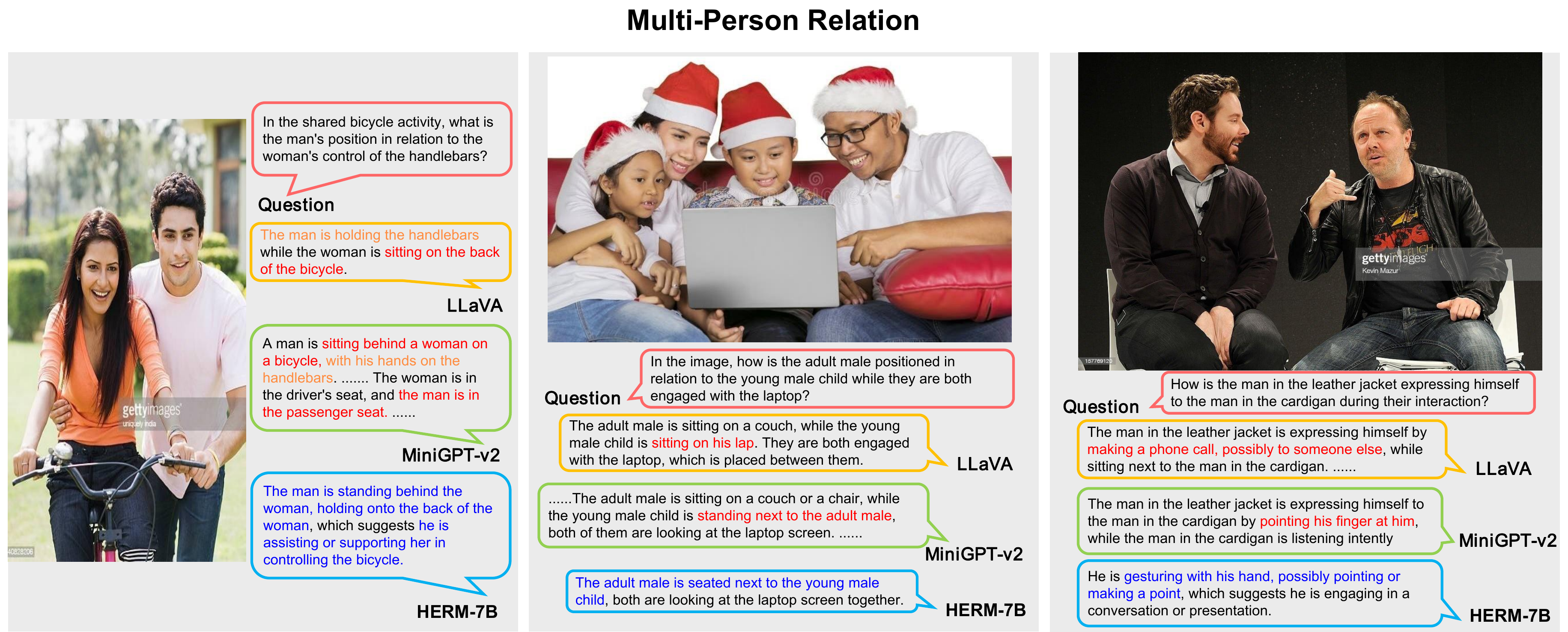}
    \vspace{-5pt}
  \caption{Evaluation examples on Multi-Person Relation Dimension.}
  \label{fig:Multi-Person-Relation}
\end{figure}

\begin{figure}[h]
  \centering
  \includegraphics[width=0.95\linewidth]{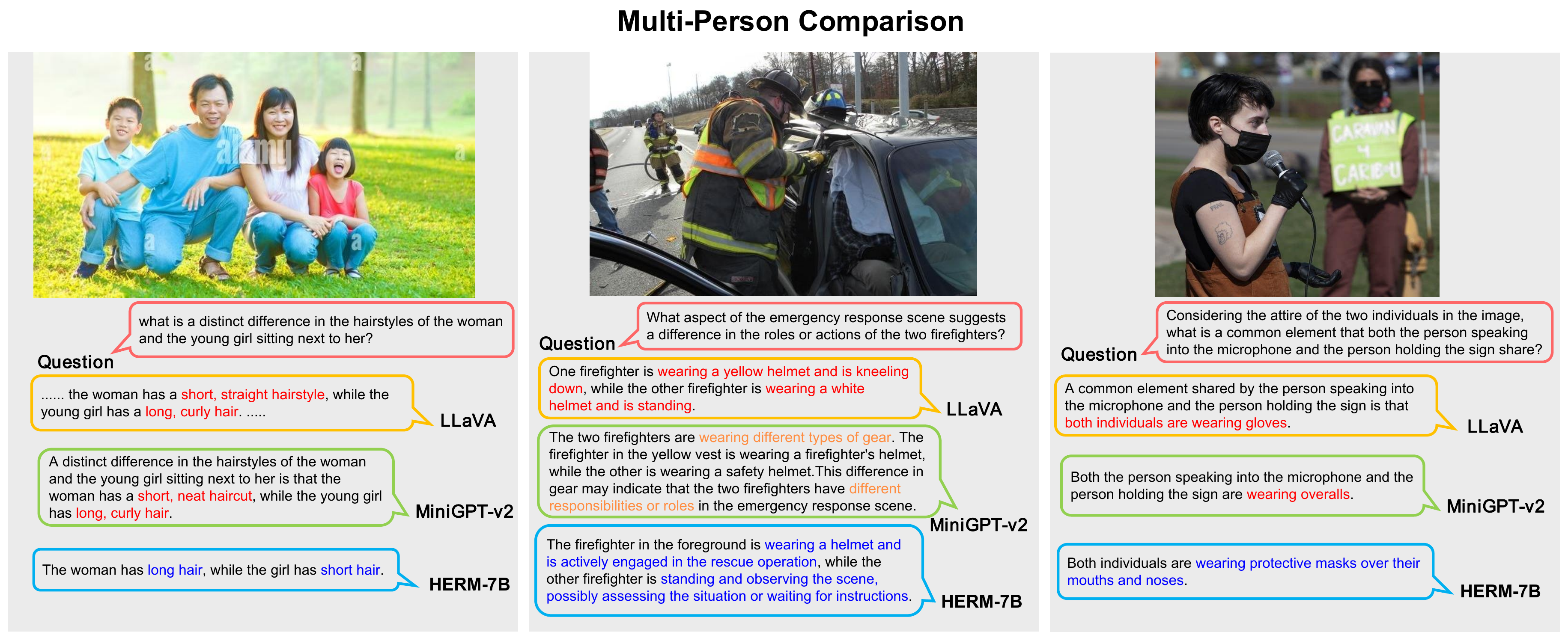}
    \vspace{-5pt}
  \caption{Evaluation examples on Multi-Person Comparison Dimension.}
  \label{fig:Multi-Person-Comparison}
\end{figure}

\begin{figure}[h]
  \centering
  \includegraphics[width=0.95\linewidth]{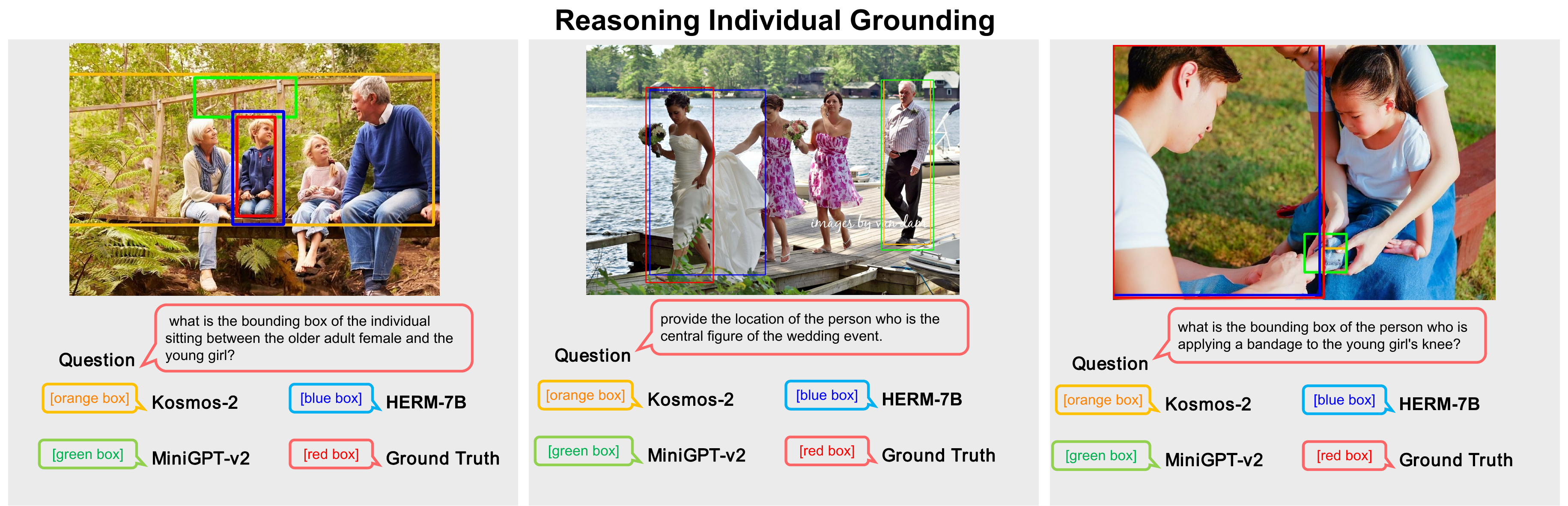}
    \vspace{-5pt}
  \caption{Evaluation examples on Reasoning Individual Grounding Dimension.}
  \label{fig:Reasoning_Individual_Grounding}
\end{figure}

In this section, 
\cref{fig:Individual_Appearance} -~\cref{fig:Reasoning_Individual_Grounding} provide additional demonstrations of qualitative examples on HERM-Bench. Each figure provides three demonstrations of a single task dimension. The models selected for comparison are LLaVA~\cite{liu2024visual}, MiniGPT-v2~\cite{chen2023minigpt} and HERM-7B. For multi-choice questions, we mark error parts in {\color{red} red}, ambiguous parts in {\color{orange} orange}, correct parts in {\color{blue} blue}. For grounding questions, we use distinct bbox color to distinguish different models' output.
As shown in these figures, our method can produce more precise human-centric understanding over existing MLLMs across various dimensions.

\bibliographystyle{splncs04}
\bibliography{main}
\end{document}